\documentclass[sigconf,screen,nonacm]{acmart}
\copyrightyear{2023}
\acmYear{2023}
\usepackage{CJKutf8}

\usepackage{array}
\usepackage{algorithm}
\usepackage{algpseudocode}

\usepackage{soul}

\algtext*{EndFor}
\algtext*{EndIf}
\algtext*{EndProcedure}
\algtext*{EndFunction}

\usepackage[utf8]{inputenc} %
\usepackage{lipsum} %

\usepackage[T1]{fontenc}

\usepackage{amsmath,amsfonts,amsthm}

\ifdefined\isoverfull
\else
\fi

\usepackage{etoolbox}

\newbool{includeappendix}
\setbool{includeappendix}{true} %
\IfFileExists{headers/config/noappendix.config}{
	\setbool{includeappendix}{false}
}{}

\newif\ifincludeappendixx
\ifbool{includeappendix}{
	\includeappendixxtrue
}{
	\includeappendixxfalse
}

\usepackage{xr} %
\usepackage{filecontents}

\ifbool{includeappendix}{}{

	\externaldocument{appendix-labels}
}

\newcommand{\eg}{\textit{e.g.,} }
\newcommand{\ie}{\textit{i.e.,} }

\definecolor{my-full-blue}{HTML}{1F77B4}

\definecolor{my-full-orange}{HTML}{FF7F0E}

\definecolor{my-full-green}{HTML}{2CA02C}

\definecolor{my-full-red}{HTML}{d62728}

\definecolor{my-full-purple}{HTML}{9467bd}

\colorlet{my-blue}{my-full-blue!30}
\colorlet{my-orange}{my-full-orange!30}
\colorlet{my-green}{my-full-green!30}
\colorlet{my-red}{my-full-red!30}
\colorlet{my-purple}{my-full-purple!30}

\usepackage{listings}

\usepackage{textcomp}

\usepackage{xcolor}

\usepackage[scaled=0.8]{beramono}

\definecolor{ckeyword}{HTML}{7F0055}
\definecolor{ccomment}{HTML}{3F7F5F}
\definecolor{cstring}{HTML}{2A0099}

\lstdefinestyle{numbers}{
	numbers=left,
	framexleftmargin=20pt,
	numberstyle=\tiny,
	firstnumber=auto,
	numbersep=1em,
	xleftmargin=2em
}

\lstdefinestyle{layout}{
	frame=none,
	captionpos=b,
}

\lstdefinestyle{comment-style}{
	morecomment=[l]//,
	morecomment=[s]{/*}{*/},
	commentstyle={\color{ccomment}\itshape},
}

\lstdefinestyle{string-style}{
	morestring=[b]",%
	morestring=[b]',%
	stringstyle={\color{cstring}},
	showstringspaces=false,%
}

\lstdefinestyle{keyword-style}{
	keywordstyle={\ttfamily\bfseries},
	morekeywords={
		function,
		constructor,
		int,
		bool,
		return,
		returns,
		uint
	},
	morekeywords = [2]{},
	keywordstyle = [2]{\text},
	sensitive=true,
}

\lstdefinestyle{input-encoding}{
	inputencoding=utf8,
	extendedchars=true,
	literate=
	{ℝ}{$\reals$}1%
	{→}{$\rightarrow$}1%
	{α}{$\alpha$}1%
	{β}{$\beta$}1%
	{λ}{$\lambda$}1%
	{θ}{$\theta$}1%
	{ϕ}{$\phi$}1%
}

\lstdefinestyle{escaping}{
	moredelim={**[is][\color{blue}]{\%}{\%}},
	escapechar=|,
	mathescape=true
}

\lstdefinestyle{default-style}{
	basicstyle=\fontencoding{T1}\ttfamily\footnotesize,
	style=numbers,
	style=layout,
	style=comment-style,
	style=string-style,
	style=keyword-style,
	style=input-encoding,
	style=escaping,
	tabsize=2,
	upquote=true
}

\lstdefinelanguage{BASIC}{
	language=C++,
	style=default-style
}[keywords,comments,strings]%

\usepackage{csvsimple}

\usepackage[capitalize]{cleveref}

\makeatletter
\renewcommand\theHALG@line{\thealgorithm.\arabic{ALG@line}}
\makeatother

\crefformat{section}{\S#2#1#3}

\crefrangeformat{section}{\S#3#1#4\crefrangeconjunction\S#5#2#6}

\crefmultiformat{section}{\S#2#1#3}{\crefpairconjunction\S#2#1#3}{\crefmiddleconjunction\S#2#1#3}{\creflastconjunction\S#2#1#3}

\newcommand{\crefrangeconjunction}{--}

\crefname{theorem}{Thm.}{Thms.}

\crefname{algorithm}{Alg.}{Algs.}
\crefname{listing}{Lst.}{listings}
\crefname{line}{Line}{Lines}
\crefname{appendix}{App.}{App.}

\newbool{reftoextendeddone}
\setbool{reftoextendeddone}{false}

\newcommand{\reftoextended}{%
	\ifbool{reftoextendeddone}{\textsuperscript{\ref{footnote:extendedversion}}}{%
		\footnote{\label{footnote:extendedversion}Available in the extended version of this paper:\\ \url{\extendedpaperlink}}%
		\setbool{reftoextendeddone}{true}%
	}%
}
\newcommand{\apprefrange}[2]{%
	\ifbool{includeappendix}{\crefrange{#1}{#2}}{App.\reftoextended{}~\hyperref[footnote:extendedversion]{\ref*{#1}}--\hyperref[footnote:extendedversion]{\ref*{#2}}}%
}
\newcommand{\appref}[1]{%
	\ifbool{includeappendix}{\cref{#1}}{App.\reftoextended{}~\hyperref[footnote:extendedversion]{\ref*{#1}}}%
}
\newcommand{\Appref}[1]{%
	\ifbool{includeappendix}{\cref{#1}}{App.\reftoextended{}~\hyperref[footnote:extendedversion]{\ref*{#1}}}%
}

\definecolor{light_red}{rgb}{0.96, 0.76, 0.76}
\definecolor{light_green}{rgb}{0.7, 0.98, 0.87}
\definecolor{light_yellow}{rgb}{0.99, 0.97, 0.37}
\definecolor{dark_green}{rgb}{0.0, 0.5, 0.0}

\newcommand{\yx}[1]{\textcolor{black}{#1}}
\usepackage{adjustbox}
\usepackage[normalem]{ulem} % use normalem to protect \emph
\newcommand{\etal}{\textit{et al}. }

\newcommand\bitzero{\bgroup\markoverwith
  {\textcolor{light_red}{\rule[-.5ex]{2pt}{2.6ex}}}\ULon}
\newcommand\bitone{\bgroup\markoverwith
  {\textcolor{light_green}{\rule[-.5ex]{2pt}{2.6ex}}}\ULon}

\usepackage{graphicx}
\usepackage{mathtools}%
\usepackage{enumitem}
\usepackage{bbm}    %
\usepackage{xspace}
\usepackage{textpos}	%
\usepackage[margin=4pt]{subcaption}	%

\usepackage{amsmath,amsfonts,bm}
\usepackage{amsthm}

\def\1{\bm{1}}

\DeclareMathAlphabet{\mathsfit}{\encodingdefault}{\sfdefault}{m}{sl}
\SetMathAlphabet{\mathsfit}{bold}{\encodingdefault}{\sfdefault}{bx}{n}

\makeatletter
\renewcommand{\paragraph}{\textbf} 
    \def\@IEEEsectpunct{.\ }

    \def\parax{\@startsection%
        {paragraph}%
        {4}%
        {0\parindent}%
        {0.6ex plus 0.1ex minus 0.1ex}%
        {0ex}%
        {\normalfont\normalsize\itshape\bfseries}%
        *%
    }%
\makeatother

\newcommand{\para}[1]{\parax{#1}.\ } %

\usepackage{booktabs}
\usepackage[normalem]{ulem}

\usepackage{stackengine}

\newenvironment{block}%
  {\vspace{1em}\list{}{\leftmargin=1em\rightmargin=1em}  \item[]  }%
	  {\endlist\vspace{1em}}

\usepackage{multirow}

\usepackage{balance}

\newcommand{\extendedpaperlink}{https://www.sri.inf.ethz.ch/publications/jovanovic2022phoenix}
\begin{document}
\title{
% Injecting Robust Authentication Watermarks into Text Written by Black-Box Language Models
Watermarking Text Generated by Black-Box Language Models
}

\author{Xi Yang}
\affiliation{%
    \institution{\small University of Science and Technology of China}\country{}
}
\email{yx9726@mail.ustc.edu.cn}
\author{Kejiang Chen}
\authornote{Corresponding authors. Our code will be available at \url{https://github.com/Kiode/Text_Watermark_Language_Models}.}
\author{Weiming Zhang}
\authornotemark[1]
\affiliation{%
    \institution{\small University of Science and Technology of China}\country{}
}
\email{{chenkj|zhangwm}@ustc.edu.cn}

% \author{Weiming Zhang}
% \authornotemark[1]
% \affiliation{%
%     \institution{University of Science and Technology of China}\country{}
% }
% \email{zhangwm@ustc.edu.cn}
\author{Chang Liu}
\author{Yuang Qi}
\affiliation{%
    \institution{\small University of Science and Technology of China}\country{}
}
\email{{hichangliu|qya7ya}@mail.ustc.edu.cn}
\author{Jie Zhang}
\affiliation{%
    \institution{\small Nanyang Technological University}\country{}
}
\email{jie_zhang@ntu.edu.sg}
\author{Han Fang}
\affiliation{%
    \institution{\small National University of Singapore}\country{}
}
\email{fanghan@nus.edu.sg}
\author{Nenghai Yu}
\affiliation{%
    \institution{\small University of Science and Technology of China}\country{}
}
\email{ynh@ustc.edu.cn}

\begin{abstract}
  Large language models now exhibit human-like skills in different fields, leading to worries about misuse like spreading misinformation and enabling academic dishonesty. Thus, detecting generated text is crucial. However, passive detection methods that train text classifiers are stuck in domain specificity and limited adversarial robustness. To achieve reliable detection, a watermark-based method was proposed for white-box language models, allowing them to embed watermarks during text generation. The method involves randomly dividing the model's vocabulary to obtain a special list and adjusting the output probability distribution to promote the selection of words in the list at each generation step. A detection algorithm aware of the special list can identify between watermarked and non-watermarked text. However, this method is not applicable in many real-world scenarios where only black-box language models are available. For instance, third-parties that develop API-based vertical applications cannot watermark text themselves because API providers only supply generated text and withhold probability distributions to shield their commercial interests. 

To allow third-parties to autonomously inject watermarks into generated text, we develop a watermarking framework for black-box language model usage scenarios. Specifically, we first define a binary encoding function to compute a random binary encoding corresponding to a word. The encodings computed for non-watermarked text conform to a Bernoulli distribution, wherein the probability of a word representing bit-1 being approximately 0.5. To inject a watermark, we alter the distribution by selectively replacing words representing bit-0 with context-based synonyms that represent bit-1. A statistical test is then used to identify the watermark. Experiments demonstrate the effectiveness of our method on both Chinese and English datasets. Furthermore, results under sentence re-translation, sentence polishing, word deletion, and synonym substitution attacks reveal that it is arduous for attackers to remove the watermark without compromising the original semantics. 

% To enable third-parties to autonomously inject watermarks into generated text, we propose a watermarking framework for black-box language model usage scenarios. Specifically, we first construct a binary encoding function to compute a random binary encoding corresponding to a word. The encodings computed for non-watermarked text conform to a Bernoulli distribution, wherein the probability of a word representing bit-1 being approximately 0.5. To inject a watermark, we alter the distribution by selectively replacing words representing bit-0 with context-based synonyms that represent bit-1. A statistical test is then used to detect the watermark.
\end{abstract}


% \ccsdesc[500]{Security and privacy~Privacy-preserving protocols}

\maketitle

\section{Introduction} \label{sec:intro}
Recent advances in large language models (LLMs) have enabled them to reach human-level proficiency across numerous professional and academic tasks \cite{instructgpt,llama,gpt4}. One of the most impressive examples is OpenAI's ChatGPT \cite{chatgpt}, which has demonstrated remarkable prowess in answering questions, composing emails, essays, and even generating code. However, this impressive ability to create human-like text with remarkable efficiency has ignited apprehension regarding the potential abuse of LLMs for malicious purposes \cite{nature_worry,nature_worry2,whatif,liebrenz2023generating}, such as phishing, disinformation campaigns, and academic dishonesty. 
% How serious is that? 
Several countries and institutions have imposed bans on ChatGPT, citing concerns about privacy breaches, ideological influences, and academic dishonesty \cite{BBC2021,newyork_ban}. Additionally, media outlets have cautioned the public regarding the possibility of misleading information generated by LLMs \cite{guardian2023}.
These growing concerns have cast a shadow on the positive applications of LLMs. Therefore, detecting and authenticating generated text becomes crucial to ensure the responsible and secure use of LLMs.
% in vital sectors such as media and education. 
\begin{figure}[t]
    \centering
    \includegraphics[width=\linewidth]{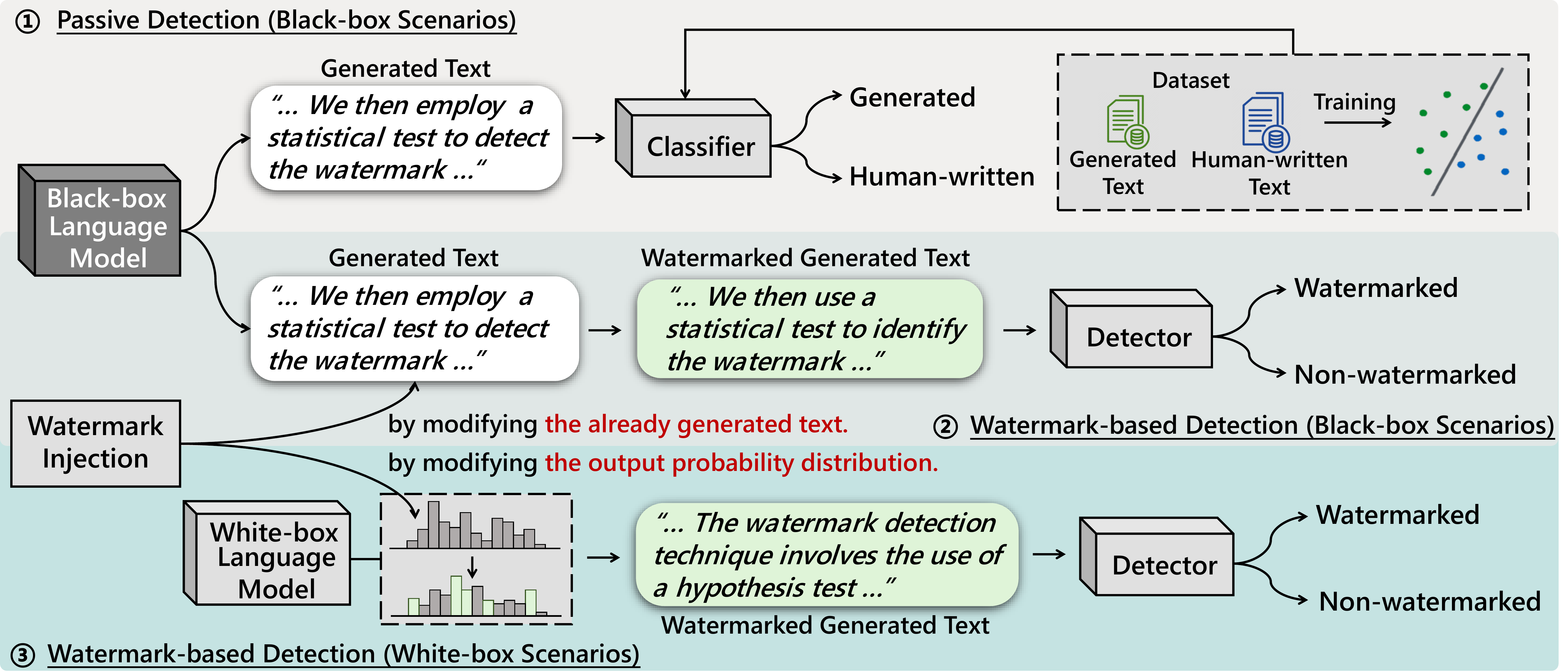}
    \caption{Flowchart of different generated text detection methods, top to bottom: Passive detection, watermark-based detection (our method) in black-box model scenarios, and watermark-based detection in white-box model scenarios.}
    \label{fig:comp}
\end{figure}

A prevalent solution is passive detection \cite{hc3,detectgpt,gptzero,openai_detect,deepfaketext,AITextDetector}, where a text classifier, usually fine-tuned on a pretrained language model like RoBERTa \cite{roberta} and GPT-2 \cite{gpt2}, is adopted to distinguish between generated and human-written text. However, these learning-based methods perform well only when the input data share a similar distribution with the training data, thereby limiting their applicability to specific domains. Moreover, as LLMs advance rapidly and human reliance on generated content grows, the line between human-written and generated text will gradually become more indistinct. For example, in the evaluations on a ``challenge set'' of English texts, OpenAI's text classifier only identifies 26\% of generated text \cite{openai_detect}.
Besides, these classifiers are vulnerable to adversarial attacks \cite{bertattack,textattack,pu2022deepfake} and are biased against non-native language writers \cite{Liang2023GPTDA}, causing more false positives and negatives. 

To achieve more reliable detection, Kirchenbauer \etal \cite{maryland} proposed a watermark-based detection method for white-box language model usage scenarios. The watermark is injected by selecting a random set of ``greenlist'' words from the model's vocabulary, and softly facilitating the generation of words in the greenlist during the sampling process. \yx{This results in a significantly increased frequency of greenlist words within the generated text, allowing for watermark detection through a statistical test.} Unlike passive detection methods, watermark-based methods do not rely on any human-written and generated text-writing features, as Figure \ref{fig:comp} shows. Besides, the statistical test-based detection is more transparent and intelligible. However, in real-world scenarios where only black-box language models are available, manipulating the probability distribution of the model's vocabulary and intervening in the generation process are not feasible, limiting the applicability of this method. 
For instance, third-parties that develop vertical applications (\eg healthcare, finance) using APIs are unable to embed watermarks into text on their own, as the APIs only provide generated text without probability distributions. 
Addressing this limitation is crucial, as the main market opportunity for LLMs is to serve as platforms for developing vertical applications \cite{ABCNews}. Furthermore, several political entities are drafting policies requiring application providers to label their generated content \cite{china,eu} , which is a prerequisite for obtaining approval to launch vertical applications.
% . Thus, compliance with regulations is critical to obtain approval for the launch of vertical applications.

% For example, third parties developing downstream services using APIs. This is because the API providers are unlikely to disclose probability distribution information to avoid exposing commercial secrets or increasing risks associated with adversarial attacks. As a result, the white-box watermarking method can not fulfill the needs of third parties seeking to embed watermarks in their generated text.
% Therefore, this method is only suitable for proprietary language models that have white-box access to the model internals.
% One possible reason why API providers do not provide probability distribution information is that it may expose their trade secrets or compromise their competitive advantage1. Providing probability distribution information may also increase the risk of adversarial attacks or reprogramming of the language models2. Moreover, probability distribution information may not be useful or interpretable for most API users who only care about the quality and efficiency of the generated text3.
% Let's read the rest of the section with the following specific example:
% \begin{block}
% 	\emph{How can an entity (AI or human) prove that this paper was written with its involvement?}
% \end{block}
% and these third parties often have the need to inject their own watermarks on their generated content.
To enable third-parties using black-box language models to autonomously watermark text for the purpose of detection or authentication,
% To address the need for watermark injection and detection in text generated from black-box language model services, 
we propose a watermarking framework for injecting watermarks into the already generated text. 
Our method begins with constructing a binary encoding function that computes a random binary representation (either bit-0 or bit-1) for a given word, based on the hash value of the word and its immediately preceding word in the text. Given that a well-designed hash function provides nearly uniformly distributed outputs, the binary encodings derived from each word in a common text are expected to approximate a Bernoulli distribution \cite{evans2011statistical} with equal probabilities of 0.5 for a word representing bit-0 and bit-1. 
Then, we inject the watermark by selectively substituting words signifying bit-0 with synonyms representing bit-1. This leads to a higher proportion of bit-1 occurrences within the binary encodings derived from the watermarked text. To maintain the original semantics during watermark insertion, we employ BERT \cite{bert} to produce context-based synonyms and introduce sentence-level and word-level similarity assessments to select high-quality synonyms. \yx{Lastly, leveraging the prior knowledge of the differences in the binary encoding distributions between watermarked and non-watermarked text}, we use a statistical test to detect the watermark in text.
% Specifically, we employ BERT \cite{bert} to generate synonym candidates for a word, and evaluate their semantic similarity to the original word using text similarity metrics. Then, a synonym sampling algorithm is designed to inject the watermark while maintaining the original semantic quality. 
% Specifically, we identify words in the original text suitable for modification and generate corresponding context-based synonym candidates that can maintain the original semantics. Then, we design a watermark-driven synonym sampling algorithm that randomly selects synonyms to replace the original words for watermark injection. Lastly, based on a statistical test, we design a detection scheme with fast and precise detection modes to identify the presence of watermarks in a given text. 
Experiments demonstrate the effectiveness of our method in injecting authentication watermarks in both Chinese and English text while maintaining the semantic integrity. Considering that in the real world, humans may post-process the text and attackers may attempt to remove the watermark by modifying the text, we evaluate the robustness of our method against sentence-level attacks (\ie re-translation and polishing) and word-level attacks (\ie word deletion and synonym substitution). The results indicate that it is difficult to remove our watermark without compromising the original semantics.
% With our watermarking framework, third-parties employing black-box language model services are able to autonomously detect or authenticate the text without interfering with the model's generation process. 
% Besides, as the human brain language system currently resembles a black-box language model, our method is also applicable for injecting watermarks into human-written texts.
\begin{block}
	\emph{By the way, the abstract of this paper contains an invisible watermark that can be identified by our watermark detector with a statistical significance level of 99\%.}
\end{block}

\para{Main Contributions} In summary, our main contributions are:
\begin{itemize}[leftmargin=12pt]
\item We present a framework for injecting authentication watermarks into text generated by black-box language models. This enables third-parties that employ black-box model services (\eg APIs) to autonomously detect or authenticate their generated content through watermarking.

\item \yx{We design a context-based synonym generation algorithm and a watermark-driven synonym sampling algorithm to achieve watermark injection without compromising the original semantics. Considering different detection time preferences, we provide a detection algorithm with two optional modes: a fast mode for quicker results and a precise mode for enhanced precision.}

\item Extensive experiments on both Chinese and English datasets showcase that our method can effectively watermark natural text while preserving the original semantics. Moreover, we simulate potential attacks (\ie re-translation, polishing, word deletion, and synonym substitution) to illustrate the difficulty in erasing the watermark without degrading the semantic quality.
% Moreover, we develop sentence-level and word-level attacks to illustrate the challenge for attackers in erasing the watermark without degrading the semantic quality.
\end{itemize}

\section{Background and related works} \label{sec:background}

% We next introduce the background necessary to present \tool. In
% \cref{ssec:background:fhe} and \cref{ssec:background:fhenn} we recap fully
% homomorphic encryption and how it is commonly used for privacy-preserving neural
% network inference. In \cref{ssec:background:robustness} and
% \cref{ssec:background:fairness} we formally introduce the two types of
% reliability guarantees considered in \tool: local robustness and individual
% fairness. Finally, in \cref{ssec:background:smoothing} we discuss randomized
% smoothing, a technique commonly used to provide such guarantees. \tool lifts
% this technique to fully homomorphic encryption to ensure predictions enjoy both
% client data privacy as well as reliability guarantees.

\subsection{Large Language Models}
\label{ssec:background:llms}
The advent of the transformer architecture \cite{Transformer} has led to a paradigm shift in natural language processing, with large language models (LLMs) human-like proficiency in various tasks \cite{zhao2023survey}. We introduce here two types of LLMs relevant to this paper, \ie autoregressive LLMs and autoencoding LLMs.
\para{Autoregressive LLMs} Autoregressive LLMs, such as GPT-3 \cite{gpt3}, generate text by predicting the next word in a sequence based on the previous words.
% In autoregressive LLMs such as GPT-3 \cite{gpt3},
The model is trained on a large corpus of text to learn the statistical patterns and relationships between words. During training, the model’s parameters are optimized to minimize the negative log-likelihood of the training data, which is equivalent to maximizing the likelihood of the target sequence given the input sequence. Mathematically, the training objective is represented as:

\begin{equation}
\mathcal{L}_{ar} = -\sum_{t=1}^{n} \log P(w_t | w_{1}, w_{2}, \dots, w_{t-1}; \theta_{ar})
\end{equation}
where 
% $\mathcal{L}_{ar}$ is the loss function, 
$w_t$ denotes the word at position $t$, and $\theta_{ar}$ represents the model parameters.
During text generation, 
% the model generates one word at a time. At each time step, 
the model samples the next word $w_t$ from the conditional probability distribution $P(w_t | w_{1}, w_{2}, \dots, w_{t-1})$ over the full vocabulary at each time step. Different sampling strategies \cite{sampling} can be used to control the trade-off between diversity and coherence in the generated text.

\para{Autoencoding LLMs} Autoencoding LLMs, such as BERT \cite{bert}, are trained with a masked language modeling (MLM) objective, which aims to predict missing words in a given context. Unlike autoregressive models that predict words sequentially, autoencoding models focus on capturing bidirectional context by simultaneously conditioning on words before and after the target word. During training, the model is presented with text where some words have been randomly masked, and the objective is to predict the original words based on their surrounding context. The training objective is to maximize the likelihood of predicting the masked words correctly based on their surrounding context:
\begin{equation}
\mathcal{L}_{ae} = -\sum_{t \in \mathcal{M}} \log P(w_t | w_{1}, w_{2}, \dots, w_{t-1}, w_{t+1}, \dots, w_n; \theta_{ae})
\end{equation}
where $\theta_{ae}$ represents the model parameters, $w_1, w_2, \dots, w_n$ are words in the text, and $\mathcal{M}$ denotes the set of masked positions. 

In BERT, each word is first tokenized and represented as a one-hot vector. This one-hot vector is then multiplied by an embedding matrix to produce the initial word embedding. The initial embedding is combined with positional and segment embeddings before being fed into the transformer encoder. The transformer encoders update the embeddings by iteratively applying self-attention and feedforward layers to capture the bidirectional context of each word. Specifically, the final hidden state corresponding to a masked word is fed into an output layer with a softmax activation function to produce a probability distribution over the vocabulary. The model predicts the masked word by selecting the word with the highest probability. The bidirectional context encoding allows BERT to excel in tasks that necessitate a deep understanding of the context, which is why we employ it to generate synonyms in our method.
% Inspired by this, we opt to use BERT for generating synonyms.
% This bidirectional context encoding allows autoencoding models like BERT to excel in tasks that require a deep understanding of the context, such as text classification, sentiment analysis, .

% \subsection{Risks of AI-generated Texts}
% The widespread use of advanced large language models (LLMs), such as ChatGPT, has raised concerns over the potential for the spread of misinformation, political manipulation, privacy breaches, and cybersecurity threats. Additionally, there are concerns about academic misconduct, such as students using LLMs to cheat and produce academically dishonest work. In light of these challenges, it is important to develop methods for detecting text generated by LLMs in order to mitigate the risks and consequences. 

\subsection{Recent Generated Text Detection Methods}
% Recently, various detection methods based on different features have been proposed.
\para{Statistical Discrepancy Detection} Several methods distinguish between generated and human-written text by identifying statistical discrepancies between them, as exemplified by two recent tools: GPTZero \cite{gptzero} and DetectGPT \cite{detectgpt}. GPTZero uses perplexity and burstiness to tell apart human-written and generated text, as language models tend to produce more predictable and consistent text based on the patterns they learned from training data, resulting in lower perplexity scores for generated text. DetectGPT exploits the negative curvature regions of a model's log probability function to identify generated text by comparing the log probability of unperturbed and perturbed text variations. However, as language models are constantly improving and becoming more sophisticated, these heuristic features struggle to achieve robustness and generalization.
% , leading to a high rate of false positives.
\para{Deep Learning-based Detection} Deep learning-based methods rely on gathering human-written and generated samples to train classifiers. Recently, OpenAI fine-tuned a GPT model for this discrimination task using a dataset comprising paired human and AI-generated texts on identical topics \cite{openai_detect}. Similarly, Guo \etal \cite{hc3} fine-tuned a text classifier based on pre-trained autoencoding LLMs (\eg RoBERTa) by collecting the Human ChatGPT Comparison Corpus (HC3). Deep learning-based methods exhibit strong performance under the training data distribution, but they are susceptible to adversarial attacks, lack interpretability, and struggle to provide reliable judgments in human-AI collaboration scenarios.
\para{Watermark-based Detection} Kirchenbauer \etal \cite{maryland} proposed the watermarking framework for white-box language models. The watermarking operates by randomly selecting a random set of "greenlist" words from the model's vocabulary and softly encouraging the use of these "greenlist" words by interfering with the sampling process at each generation step. The watermark can be detected by testing the following null hypothesis, 
\begin{block}
	\emph{${H_0}$: The text sequence is generated with no knowledge of the selection rule of "greenlist" words.}
\end{block}
If the null hypothesis is rejected, it can be concluded that the text was generated by the given model. 
This method is suitable for model owners who have access to the model’s output probability distribution and can interfere with the sampling process. However, it is not feasible for third parties who develop vertical applications using black-box language model services (\eg APIs) and do not have access to the model’s internals, even though they also have a need to embed watermarks in text generated from them.  

\subsection{Multi-Bit Text Watermarking Methods} \label{ssec:multibit_watermark}
Traditional text watermarking tries to embed a multi-bit watermark within the text, aiming to facilitate tracing the text provenance. Abdelnabi and Fritz \cite{awt} proposed a transformer-based encoder-decoder network, named AWT, that can embed fixed-length watermark information in English text. The network learns to replace inconspicuous words (\eg prepositions, conjunctions, and symbols) with similar alternatives to encode information, resulting in a robust watermark that can be extracted even if some words are altered, provided the inconspicuous words remain intact. However, although the authors introduced sentence embedding constraints to maintain the semantic quality of the watermarked text, the network did not genuinely focus on semantic quality. Instead, it learned to modify words with minimal impact on sentence embedding (such as prepositions and symbols), leading to watermarked text with numerous grammatical errors and distortions. Additionally, sentence-level attacks (\eg polishing, rearranging sentence order) can result in the disorder and length changes of the extracted bits, causing the watermark bits to lose synchronization.

Yang \etal \cite{yangxi} proposed a synonym substitution algorithm for embedding a multi-bit watermark within a given text. This method offers superior semantic quality compared to AWT. However, it requires the watermark embedder and extractor to locate the same words and generate identical synonyms to achieve successful watermarking. Moreover, their watermarking algorithm is highly sensitive to context changes, any slight alteration of the context may cause the watermark bits to be desynchronized and unextractable.

\section{Motivation} \label{sec:motivation}
\yx{Our objective is to design a framework that enables text generation service providers to perform watermark injection and detection in the text generated from black-box language models (where only model outputs are observable, rather than parameters or internal computations).
In this paper, we primarily consider two entities: the attacker and the text generation service provider. The attacker seeks to exploit the generated text for malicious purposes, while the service provider aims to detect or authenticate the text by verifying the presence of a watermark, thus helping to mitigate the abuse of its services. The attacker may post-process the generated text without compromising the original semantics. But they will not completely rewrite the text, as doing so contradicts the purpose of using the text generation service. Therefore, the watermarking framework should have the following properties:
}
% \yx{Our goal is to propose a watermarking framework to inject watermarks into text generated from black-box language models.} For practical scenarios, the watermarking framework should have the following properties:
\begin{itemize}[leftmargin=12pt]
\item \textbf{Fidelity}: The injection of a watermark should not affect the original semantic information.

\item \textbf{Robustness}: Attackers should not be able to erase the watermark without compromising the original semantic information.

\item \textbf{Generality}: The watermarking framework should work for text written in different languages and covering different topics.
\end{itemize}

\section{Our Method} \label{sec:method} 
\begin{figure}[t]
    \centering
    \includegraphics[width=\linewidth]{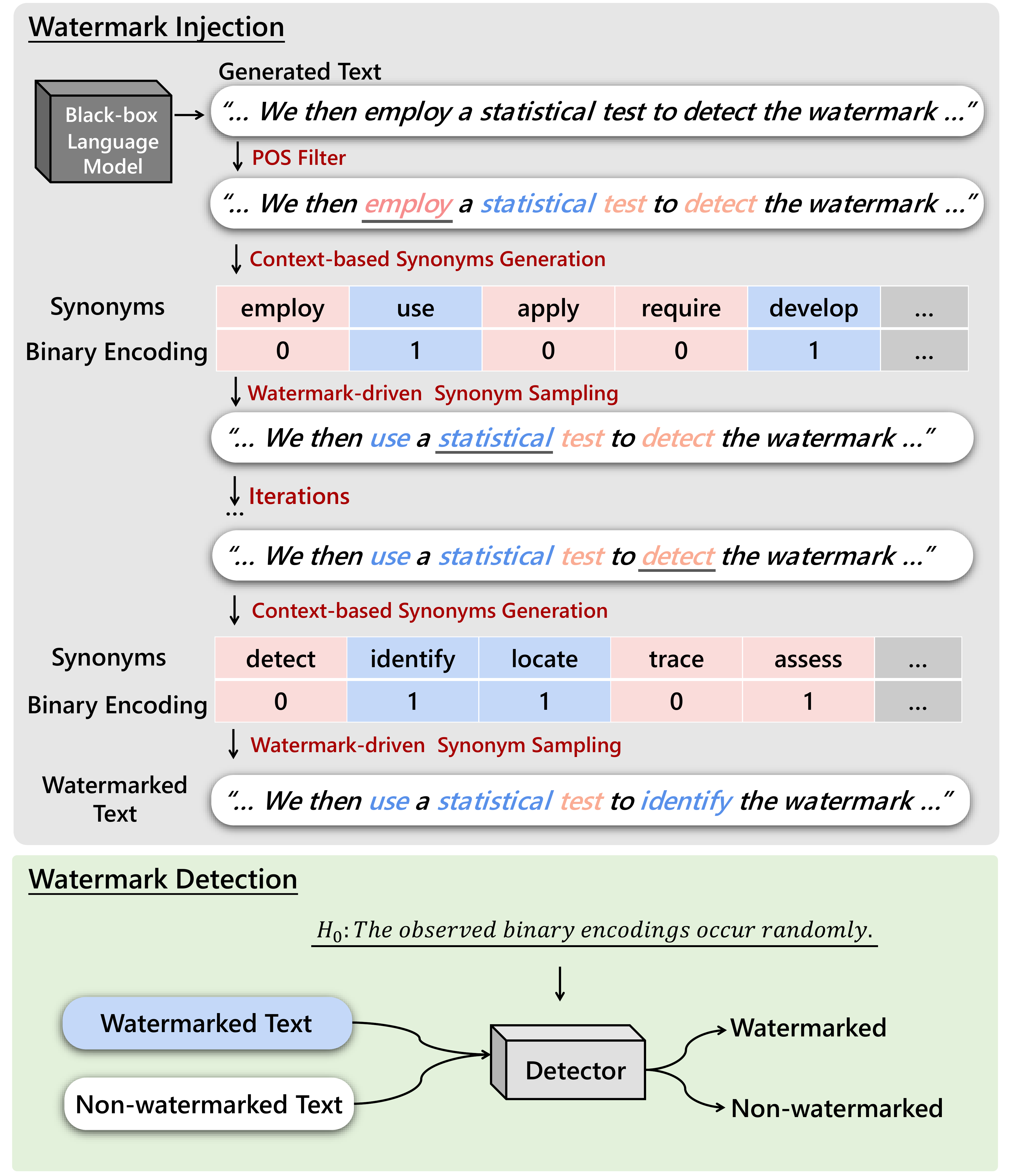}
    \caption{The proposed watermarking framework.}
    % , depicting the process of injecting a watermark into text generated by a black-box language model and the corresponding detection process.}
    \label{fig:workflow}
\end{figure}

In this section, we will elaborate the proposed watermarking framework. As illustrated in Figure \ref{fig:workflow}, once we acquire the original generated text from a black-box language model, we selectively and sequentially replace words with synonyms to inject the watermark.
% specific words that could potentially be replaced by synonyms and attempt to modify these words sequentially to inject the watermark. 
Specifically, we first construct a binary encoding function that computes a random binary representation (either bit-0 or bit-1) for a given word. This function possesses a notable property: in a non-watermarked text, the number of words representing bit-0 and bit-1 are nearly balanced. Then,
for each selected word, we first generate its context-based synonym candidates and compute the random binary encoding carried by each candidate. 
Then, we develop a watermark-driven synonym sampling algorithm to encourage the selection of candidates representing bit-1 to inject the watermark.
The injected watermark results in a relatively higher proportion of words representing bit-1. Therefore, we can employ a statistical test to detect the presence of a watermark. Subsequently, we present a comprehensive explanation of the binary encoding function and the watermarking process. 

\subsection{Binary Encoding Function} \label{sec:Design} 
Here, we design the binary encodings to express watermark information within the text.
Let $w_i$ denotes the $i$-th word in the text, and $h(\cdot)$ represents the string hash function. We utilize the combined hash of the current word and its preceding word as a seed for generating a random binary value corresponding to $w_i$. \yx{By including the preceding word, we ensure that a word demonstrates variability in expressing bit-0 and bit-1 under different contexts.} This can be formalized as follows:
% \begin{equation}
% s_i = h(t_i) \oplus h(t_{i-1}), \quad i = 2, \ldots, n
% \end{equation}
\begin{equation}
b_i = \text{RandomBinary}(h(w_i) \oplus h(w_{i-1})), \quad i = 2, \ldots, n
\label{eq:randombinary}
\end{equation}
where $\oplus$ denotes the bitwise XOR operation, and $b_i$ represents the binary encoding corresponding to $w_i$. The function $\text{RandomBinary}(x)$ produces a random bit based on the input seed.

\yx{Owing to the near-uniform nature of the hash function, the original text should exhibit a roughly equal distribution of words representing bit-0 and bit-1. Building on this, we propose to inject the watermark by altering the distribution, raising the proportion of words representing bit-1. Then, we can determine whether the text contains a watermark using a statistical testing method.}

% As shown in Figure \ref{fig:workflow}, a trusted third-party service provider obtains the raw text from a black-box large language model. The tokens in the text that are suitable for modification according to certain rules will be , such as the red and blue tokens in the Figure \ref{fig:workflow}, and generate synonym candidates based on the context, starting from the first token that is suitable for modification in a sequential order. Then, we calculate the replaceable score of each candidate token in the current context. In addition, we use the hash of the current token and its previous token as a random number seed to compute the random bits (0 or 1) for each candidate token under that condition. After that, we further filter the candidate tokens, i.e., remove all tokens with random bits of 0 and tokens with replaceable scores below the pre-set threshold. if there are still candidate tokens that meet the requirements, the original token is replaced with the token with the highest score. otherwise, the original token is not modified.

% Since the hash function is close to uniform, then the number of tokens with random bits of 0 and 1 in the original text should be close to uniform before we perform random synonym sampling. And the watermarked text after random synonym sampling will have more tokens with bit 1. In this way, we can detect whether a given text contains a watermark with a statistical test. This can thus be used to detect or authenticate whether the text contains content generated by that third-party downstream service provider.
 
\subsection{Watermark Injection} \label{ssec:injection}
The watermark injection begins with the second word of the input text and proceeds sequentially to the last word. Specifically, for the $i$-th word $w_i$ in text, we first compute its part-of-speech (POS), which is a linguistic category that refers to the syntactic role of a word in a sentence. If $w_i$ fails to pass through our POS filter, indicating that it belongs to a category that is unsuitable for substitution, we skip this word. If $w_i$ passes the POS filter and its corresponding binary encoding is bit-0, we generate synonym candidates for $w_i$ and replace it with a selected synonym that represents bit-1 using our watermark-driven synonym sampling algorithm.
% Next we introduce these modules in order.
\para{POS Filter} 
To assess whether a word is eligible for substitution, we employ language-specific exclusion lists, which are customized to accommodate the unique features of various languages and contexts. For English, our exclusion list encompasses pronouns, prepositions, conjunctions, proper nouns, punctuation marks, quantifiers, personal names, place names, and other proprietary terms. For Chinese, the exclusion list is composed of auxiliary words, proper nouns, punctuation marks, quantifiers, personal names, place names, and other proprietary terms. The exclusion list can be customized to accommodate specific needs and situations.
\para{Context-based Synonym Generation}
Since BERT's pre-training task involves predicting masked words within a text, it is well-suited for synonym generation. 
However, directly masking the target word will lose the information conveyed by the word itself, causing BERT to generate less suitable candidates for synonym substitution. To enable BERT to leverage both the contextual information and the target word's information when generating synonyms, inspired by Zhou \etal \cite{bert-ls}, we apply random dropout to the word embedding of the target word, resulting in a partially masked word.
% However, directly masking the target word may result in losing the information conveyed by the word itself. To allow BERT to obtain both contextual information and the current word's information, inspired by Zhou \etal \cite{bert-ls}, we perform random dropout on the word embedding of the current word, so that it is partially masked rather than completely masked.
Given the original word $w_i$ in text $\textbf{T}=(w_1,w_2,...,w_i,...w_n)$, we apply random dropout to the word embedding of $w_i$ to create a partially masked version, $\tilde{w}_i$. We then feed the partially masked embeddings into BERT to predict the initial set of synonym candidates, denoted as $C=\{s_1,s_2,...,s_K\}$, representing top-$K$ words predicted by BERT.

Nonetheless, since the BERT model is trained unsupervised on a large-scale corpus, it can only estimate the statistical similarity between two words (\ie the likelihood of co-occurring in the same context). As a result, it might consider antonyms as `similar', since they frequently appear in similar contexts and share similar syntactic structures. Thus, it is essential to further evaluate the semantic similarity between words in $C$ and the original word $w_i$.

We adopt three metrics to evaluate semantic similarity, namely sentence embedding similarity ($S_{\text{sent}}$), global word embedding similarity ($S_{\text{global}}$), and contextualized word embedding similarity ($S_{\text{context}}$).
Let $\mathbf{T'}$ denote the text after replacing $w_i$ with a synonym $s$ from $C$.
We use the RoBERTa model \cite{roberta}, fine-tuned on the Multi-Genre Natural Language Inference (MNLI) corpus \cite{N18}, to obtain sentence embeddings for the original text (\(\text{SentEmb}(\mathbf{T}\))) and the text after replacement (\(\text{SentEmb}(\mathbf{T'}\))). We then calculate the cosine distance between these two sentence embeddings:
\begin{equation}
S_{\text{sent}} = \cos(\text{SentEmb}(\mathbf{T}),\text{SentEmb}(\mathbf{T^\prime}))
\end{equation}

To obtain the global word embeddings, we consult the open-source Word-to-Vec models like GloVe \cite{glove}. The similarity between the global word embeddings of the candidate word and the original word can be expressed as:
\begin{equation}
S_{\text{global}} = \cos(\text{w2v}(w_i), \text{w2v}(s))
\end{equation}
where $\text{w2v}(\cdot)$ denotes the use of the Word-to-Vec model to obtain the embedding of the input word and $s$ means the synonym.

To compute the contextualized word embedding similarity using BERT, we denote the contextualized representation of a word $x$ at the $l$-th layer of BERT as ${f}^l(x,c)$. Here, $c$ refers to the context in which $x$ appears. 
\begin{equation}
S_{\text{context}} = \frac{1}{L} \sum_{l=1}^{L} \cos({f}^l(w_i, c_{T}), {f}^l(s, c_{T^\prime}))
\end{equation}
We use the last 8 hidden layers ($L=8$) of BERT for computing the contextualized word embedding similarity, considering that different layers of BERT can attend to different dimensions of semantic features \cite{wada2022unsupervised}.
To provide a more comprehensive measure of word-level similarity, we calculate the weighted average of $S_{\text{global}}$ and $S_{\text{context}}$:
\begin{equation}
S_{\text{word}} = \lambda S_{\text{context}} + (1 - \lambda) S_{\text{global}}
\end{equation}
where $\lambda$ is the relative weight, with value ranging between 0 and 1.

Then, we further filter the candidates in $C$ according to their $S_{\text{sent}}$ and $S_{\text{word}}$ scores. Specifically, we set a sentence-level similarity threshold ($\tau_{sent}$) and a word-level similarity threshold ($\tau_{word}$). 
Given candidate set $C=\{s_1,s_2,...,s_K\}$, sentence-level similarity score $S_{\text{sent}}$, and word-level similarity score $S_{\text{word}}$, the filtered candidate set $C'$ is:
\begin{equation}
C' = \{s \in C \mid S_{\text{sent}}(s, w_i) \geq \tau_{\text{sent}} \text{ and } S_{\text{word}}(s, w_i) \geq \tau_{\text{word}}\}
\end{equation}
Here, $s$ is the synonym candidate, $w_i$ is the original word. Following this, we design a synonym sampling algorithm that utilizes the synonyms in $C'$ to inject a watermark into text $\textbf{T}$.
\para{Watermark-Driven Synonym Sampling} For each candidate $s_k'$ in the filtered set $C'=\{s_1',s_2',...,s_{K'}'\}$, we compute the binary encoding represented by $s_k'$ in the current text:
\begin{equation} \label{eq:bk}
b_k = \text{RandomBinary}(h(s_k') \oplus h(w_{i-1})), \quad k = 2, \ldots, K'
\end{equation}
Here, $b_k$ is the binary encoding, $h(\cdot)$ is a hash function, and $w_{i-1}$ is the preceding word in the text. Then, we select the candidate with a binary encoding of bit-1 and the highest $S_{\text{word}}$ to replace $w_i$. Let the selected candidate be $s_{selected}$. Then, we have:
\begin{equation}
s_{selected} = \arg\max_{s_k \in C'} \{S_{\text{word}}(s_k, w_i) \mid b_k = 1\}
\end{equation}
We achieve watermark injection at this step by replacing $w_i$ with $s_{selected}$. Then, we proceed to the next word, $w_{i+1}$, and perform the same watermark injection operation, iterating until the last word. In Algorithm \ref{alg:watermark_injection}, we provide the complete watermark injection process.

\begin{algorithm}[t]
\small
\caption{Watermark Injection}\label{alg:watermark_injection}
\begin{algorithmic}[1]
\Procedure{WatermarkInjection}{$\mathbf{T}$} \Comment{$\mathbf{T}=\{w_1,w_2,...,w_n\}$}
    % \State $n \gets$ Length($\mathbf{T}$)
    \For{$i \in \{2, 3, \ldots, n\}$}
        \State $b_i \gets \text{RandomBinary}(h(w_i) \oplus h(w_{i-1}))$
        \If{POSFilter($w_i$) and $b_i==0$}
            \State $C \gets$ SynonymsGeneration($\mathbf{T}, w_i$)
            \State $C' \gets$ FilterCandidates($\mathbf{T}, C, w_i$)
            \State $s_{selected} \gets$ SynonymSampling($\mathbf{T}, C', w_i$)
            \State Replace $w_i$ with $s_{selected}$ in $\mathbf{T}$
        \EndIf
    \EndFor
\EndProcedure

\Function{POSFilter}{$w_i$}
    \If{$POS(w_i)$ is in $ExclusionList$}
        \State \Return False
    \Else
        \State \Return True
    \EndIf
\EndFunction

\Function{SynonymsGeneration}{$\mathbf{T}, w_i$}
    \State Partially mask $w_i$ as $\tilde{w}_i$
    \State Input $\tilde{w}_i$ with context to BERT and predict candidate words $C$
    \State \Return $C$
\EndFunction

\Function{FilterCandidates}{$\mathbf{T}, C, w_i$}
    \State Initialize an empty set $C'$
    \For{each $s \in C$}
        \State Replace $w_i$ with $s$ to get $\mathbf{T'}$
        % \State Calculate sentence embeddings: $\mathbf{SentEmb}(\mathbf{T})$, $\mathbf{SentEmb}(\mathbf{T'})$
        \State $S_{\text{sent}} \gets \cos(\text{SentEmb}(\mathbf{T}),\text{SentEmb}(\mathbf{T'}))$
        \State $S_{\text{global}} \gets \cos(\text{w2v}(w_i), \text{w2v}(s))$
        % \State Compute the contextualized representations $\mathbf{f}^l(t_i, c_{T})$ and $\mathbf{f}^l(w, c_{T'})$ for the last 8 hidden layers
        \State $S_{\text{context}} \gets \frac{1}{L} \sum_{l=1}^{L} \cos({f}^l(w_i, c_{T}), {f}^l(s, c_{T'}))$
        \State $S_{\text{word}} \gets \lambda S_{\text{context}} + (1 - \lambda) S_{\text{global}}$
        \If{$S_{\text{sent}} \geq \tau_{\text{sent}}$ and $S_{\text{word}} \geq \tau_{\text{word}}$}
            \State Append $s$ to $C'$
        \EndIf
    \EndFor
    \State \Return $C'$
\EndFunction

\Function{SynonymSampling}{$\mathbf{T}, C', w_i$}
    \For{each $s_k' \in C'$}
        \State Compute $b_k$ using Eq. (\ref{eq:bk})
    \EndFor
    \State $s_{selected} \gets \arg\max_{s_k' \in C'} \{S_{\text{word}}(s_k', w_i) \mid b_k = 1\}$
    \State \Return $s_{selected}$
\EndFunction
\end{algorithmic}
\end{algorithm}

\subsection{Watermark Detection}
As described in \cref{sec:Design}, for each word in the non-watermarked text, the probability of representing bit-0 and bit-1 is nearly 0.5. During watermark injection, we employ the synonym sampling algorithm to increase the occurrence of words representing bit-1. Thus, watermark detection can be accomplished by examining the following null hypothesis:
\begin{block}
	\emph{${H_0}$: The observed binary encodings occur randomly.}
\end{block}
To verify the null hypothesis ${H_0}$, we calculate the following test statistic:
\begin{equation} \label{eq:z}
Z = \frac{(\hat{p} - p_0)}{\sqrt{\frac{p_0(1-p_0)}{N}}}
\end{equation}
where $\hat{p}$ is the proportion of words representing bit-1, $p_0 = 0.5$ represents the expected proportion under the null hypothesis (\ie random binary encodings), and $N$ is the total number of binary encodings derived from the text.
We then compare the test statistic $Z$ with the critical value $Z_{\alpha}$ corresponding to the chosen significance level $\alpha$. The significance level, denoted by $\alpha$, is the probability of rejecting the null hypothesis when it is true, thereby determining the threshold for a statistically significant result. If $Z > Z_{\alpha}$, we reject the null hypothesis and conclude that the observed binary encodings are significantly different from random encodings, indicating the presence of a watermark. 

We offer two optional watermark detection modes, called fast detection and precise detection. Fast detection simply computes the binary encodings for words passing the POS filter and then conducts the hypothesis test. Precise detection further selects words highly likely to carry watermark information before performing the hypothesis test, leading to a more accurate detection scope. The pseudocode for both fast and precise detection modes can be found together in Algorithm \ref{alg:detection}. For a more intuitive understanding, we also provide examples of each mode in Table \ref{tab:example_detection}.
\para{Fast Detection}
For the text under inspection, we begin with the second word and assess whether its POS can pass our POS filter. If it fails, we skip this word; otherwise, we compute its binary encoding and continue the operation iteratively until the last word. After acquiring the binary encodings, we calculate the $Z$-score according to Eq.(\ref{eq:z}) to determine if the text contains a watermark.
% We can then compare the test statistic $Z$ with the critical value $Z_{\alpha}$ corresponding to the chosen significance level $\alpha$. If $Z > Z_{\alpha}$, we reject the null hypothesis and conclude that the observed watermark pattern is significantly different from a random pattern, indicating the presence of a watermark.

\begin{table*}[htbp]
  \small
  \centering
  \caption{Examples of watermark detection. 
    Red-highlighted words represent bit-0 and green ones bit-1; underlining shows the scope of precise detection. The $p$-value indicates the likelihood of the text not containing a watermark.}
    \vspace{-6pt}
    \begin{tabular}{c>{\raggedright\arraybackslash}p{47em}cc}
    \toprule[1.5pt]
       & \multirow{2}{*}{Text Content} & \multicolumn{2}{c}{$p$-value} \\
       \cmidrule(r){3-4}
       & & Fast & Precise \\
    \hline
    \multirow{5}{*}{\footnotesize{Original}} &
    Flocking is a \bitzero{\ul{type}} of \bitzero{{coordinated}} \bitzero{\ul{group} \ul{behavior}} that is \bitzero{{exhibited}} by \bitzero{{animals}} of \bitzero{\ul{various} \ul{species}}, \bitzero{\ul{including}} \bitzero{{birds}}, \bitzero{{fish}}, and \bitzero{{insects}}. It is \bitone{{characterized}} by the \bitone{\ul{ability}} of the \bitzero{{animals}} to \bitone{\ul{move}} \bitone{{together}} in a \bitzero{{coordinated}} and cohesive \bitzero{\ul{manner}}, as if they were a \bitzero{{single}} \bitone{{entity}}. Flocking \bitone{\ul{behavior}} is \bitzero{\ul{thought}} to have \bitzero{\ul{evolved}} as a \bitone{{way}} for \bitzero{{animals}} to \bitzero{\ul{increase}} their \bitzero{\ul{chances}} of \bitone{{survival}} by \bitzero{{working together}} as a \bitone{{group}}. For \bitzero{\ul{example}}, flocking \bitone{{birds}} \bitzero{\ul{may}} be \bitzero{\ul{able}} to \bitone{\ul{locate}} \bitzero{{food}} more \bitone{\ul{efficiently}} or \bitone{\ul{defend}} themselves against predators more \bitone{\ul{effectively}} when they \bitzero{{work}} \bitone{{together}}.
    & \multirow{5}{*}{0.9933} & \multirow{5}{*}{0.9646} 
    \vspace{0.5pt}
    \\
    \hline
    \multirow{5}{*}{\footnotesize{Watermarked}} & 
    Flocking is a \bitone{\ul{kind}} of \bitzero{coordinated} \bitone{\ul{team} \ul{behavior}} that is \bitzero{exhibited} by \bitzero{animals} of \bitone{\ul{several} \ul{species}}, \bitone{\ul{notably}} \bitzero{birds}, \bitzero{fish}, and \bitzero{insects}. It is \bitone{characterized} by the \bitone{ability} of the \bitzero{animals} to \bitone{\ul{move}} \bitone{together} in a \bitzero{coordinated} and cohesive
    
    \bitone{\ul{way}}, as if they were a \bitzero{single} entity. Flocking \bitone{\ul{behavior}} is \bitone{\ul{believe}} to have \bitzero{\ul{evolved}} as a \bitone{way} for \bitzero{animals} to 
    
    \bitone{\ul{raise}} their \bitone{\ul{likelihood}} of survival by \bitzero{working together} as a \bitone{\ul{group}}. For \bitone{\ul{instance}}, flocking \bitone{birds} \bitone{\ul{could}} be \bitzero{able} to \bitone{\ul{locate}}
    
    \bitone{\ul{nutrition}} more \bitone{\ul{efficiently}} or \bitone{\ul{defend}} themselves against predators more \bitone{\ul{effectively}} when they \bitzero{work} \bitone{together}.
    & \multirow{5}{*}{0.0342} & \multirow{5}{*}{0.00004} 
    \vspace{0.5pt}
    \\
    \hline
    \multirow{3}{*}{\footnotesize{Original}} & 
    \begin{CJK*}{UTF8}{gbsn}\mbox{当你\bitone{\underline{想向}}中国\bitzero{\underline{女孩子}}\bitone{\underline{说出}}第一句\bitzero{话时}，你\bitone{\underline{应该}}\bitone{先}\bitzero{\underline{考虑}}一些\bitzero{\underline{基本}}的\bitone{\underline{礼貌}}，\bitzero{\underline{例如}}\bitone{\underline{问}}她的\bitzero{\underline{名字}}\bitone{\underline{或者}}\bitzero{\underline{问}}她\bitone{\underline{是否}}\bitzero{愿意和}}

    \mbox{你\bitzero{\underline{交谈}}。
    你\bitone{也可以}\bitzero{\underline{先}}\bitzero{说出}你自己的\bitzero{\underline{名字}}，\bitone{\underline{然后}}\bitone{\underline{问}}她\bitone{\underline{是否}}\bitzero{\underline{愿意}}\bitzero{认识}你。你\bitone{可以说}："\bitone{\underline{你好}}，我\bitzero{叫}XXX，你\bitone{\underline{叫}}}
    
    \mbox{什么\bitone{\underline{名字}}？你\bitzero{\underline{愿意}}\bitzero{和}我\bitone{\underline{聊}}一\bitone{聊}吗？" \bitzero{\underline{如果}}你\bitzero{暗恋中}的\bitzero{\underline{女孩子}}\bitone{是}你的\bitzero{\underline{朋友}}，你\bitone{可以}\bitzero{\underline{先}}\bitone{\underline{尝试}}\bitone{着和}她更多地\bitzero{交流}}
    \end{CJK*} 
    % \begin{CJK*}{UTF8}{gbsn}\mbox{当你\underline{\bitone{想向}}中国\underline{\bitzero{女孩子}\bitone{说出}}第一句\bitzero{话时}，你\underline{\bitone{应该}}\bitone{先}\underline{\bitzero{考虑}}一些\underline{\bitzero{基本}}的\underline{\bitone{礼貌}}，\underline{\bitzero{例如}}\underline{\bitone{问}}她的\underline{\bitzero{名字}\bitone{或者}\bitzero{问}}她\underline{\bitone{是否}}\bitzero{愿意和}}
    % \mbox{你\underline{\bitzero{交谈}}。
    % 你\bitone{也可以}\underline{\bitzero{先}}\bitzero{说出}你自己的\underline{\bitzero{名字}}，\underline{\bitone{然后}}\underline{\bitone{问}}她\underline{\bitone{是否}}\underline{\bitzero{愿意}}\bitzero{认识}你。你\bitone{可以说}："\underline{\bitone{你好}}\underline{\bitzero{叫}}XXX，你\underline{\bitone{叫}}什么}
    % \mbox{\underline{\bitone{名字}}？你\underline{\bitzero{愿意}}\bitzero{和}
    % 我\underline{\bitone{聊}}一\bitone{聊}吗？" \underline{\bitzero{如果}}你\bitzero{暗恋中}的\underline{\bitzero{女孩子}}\bitone{是}你的\underline{\bitzero{朋友}}，你\bitone{可以}\underline{\bitzero{先}}\underline{\bitone{尝试}}\bitone{着和}她更多地\bitzero{交流}}
    % ，\underline{\bitzero{看看}}}
    % 她\underline{\bitzero{对}}你\underline{\bitzero{有没有}\bitone{兴趣}}。
    % 如果你\bitone{想向}她\bitone{表白}，你\bitone{可以说}："我\bitzero{很喜欢}你，我\bitzero{希望}我们\bitone{能够}\bitzero{在}一起。你\bitzero{愿意跟}我\bitone{约会}吗？"
    % \end{CJK*} 
    & \multirow{3}{*}{0.4443} & \multirow{3}{*}{0.4287}
    \vspace{1pt}
    \\ 
    \hline
    \multirow{3}{*}{{\footnotesize{Watermarked}}} & \begin{CJK*}{UTF8}{gbsn} \mbox{当你\bitone{\underline{想向}}中国\bitone{\underline{妹子}}\bitone{\underline{说出}}第一句\bitzero{话时}，你\bitone{\underline{应该}}\bitone{先}\bitzero{\underline{考虑}}一些\bitzero{\underline{基本}}的\bitone{礼貌}，\bitzero{\underline{例如}\bitone{\underline{问}}}她的\bitzero{\underline{名字}}\bitone{\underline{或者}}\bitone{询问}她\bitone{\underline{是否}}\bitzero{愿意和}}
    
    \mbox{你\bitzero{\underline{交谈}}。你\bitone{也可以}\bitone{\underline{首先}}\bitzero{\underline{说出}}你自己的\bitzero{\underline{名字}}。\bitone{然后}\bitone{\underline{问}}她\bitone{\underline{是否}\bitzero{\underline{愿意}}}\bitzero{认识}你。你\bitone{可以说}： "\bitone{\underline{你好}}，我\bitone{\underline{叫做}}XXX,你\bitone{\underline{叫}}}

    \mbox{什么\bitone{\underline{名字}}？你\bitzero{\underline{愿意}}\bitzero{和}我\bitone{\underline{聊}}一\bitone{\underline{聊}}吗？ "\bitone{\underline{假如}}你\bitzero{暗恋中}的\bitone{\underline{妹子}}是你的\bitzero{\underline{朋友}}，你\bitone{可以}\bitone{\underline{首先尝试}}\bitone{着和}她更多地\bitzero{交流}}
    % ，}
    % \underline{\bitone{看}}她\bitzero{对}你\underline{\bitone{有无}}\underline{\bitzero{兴趣}}。
    % 如果你想向她表白，你可以说： " 我很喜欢你，我期望我们能够在一起。你愿意跟我约会吗？"
    \end{CJK*} & \multirow{3}{*}{0.0316} & \multirow{3}{*}{0.0045}
    \\
    \bottomrule[1.5pt]
    \end{tabular}%
  \label{tab:example_detection}%
\end{table*}
\para{Precise Detection}
An enhanced detection can be devised by leveraging our prior knowledge of the watermark injection. The fast detection compute binary encodings from all words passing through the POS filter, potentially including words lacking high-quality synonyms. This could lead to the inclusion of words, which were not replaced during the watermark injection and represent bit-0, in the hypothesis test of fast detection. Therefore, we can improve detection performance by computing binary encodings and performing hypothesis test only for words that are likely to have high-quality synonyms.
Specifically, for each word that passes through the POS filter, we generate its synonym candidates with the same process in watermark injection (\cref{ssec:injection}). If the candidate set $C'$ is empty, we assume that the word is less likely to be modified during the watermark injection and exclude it from the test scope. Otherwise, we compute its binary encoding and then perform the same operation on the next word until the last word. After computing the binary encodings for all these words, we calculate the $Z$-score to determine if the text contains a watermark.

The precise detection offers more accurate watermark detection, but it comes with a higher computational cost due to the need to generate synonyms. Users can choose the optimal detection mode based on their preferred detection time.

\begin{algorithm}[t]
\small
\caption{Watermark Detection}\label{alg:detection}
\begin{algorithmic}[1]
\Procedure{FastDetection}{$\mathbf{T}$} \Comment{$\mathbf{T}=\{w_1,w_2,...,w_n\}$}
    \State Initialize binary encoding count $N \gets 0$
    \State Initialize bit-1 encoding count $c \gets 0$
    \For{$i \in \{2, 3, \ldots, n\}$}
        \If{POSFilter($w_i$)}
            \State Compute $b_k$ using Eq. (\ref{eq:bk})
            \State $N \gets N + 1$
            \If{$b_k = 1$}
                \State $c \gets c + 1$
            \EndIf
        \EndIf
    \EndFor
    \State Compute test statistic $Z$ using Eq. (\ref{eq:z})
    \State Compare $Z$ with the critical value $Z_{\alpha}$
    \State \Return watermark presence decision
\EndProcedure
\Procedure{PreciseDetection}{$\mathbf{T}$}
    \State Initialize binary encoding count $N \gets 0$
    \State Initialize bit-1 encoding count $c \gets 0$
    \For{$i \in \{2, 3, \ldots, n\}$}
        \If{POSFilter($w_i$)}
            \State $C \gets$ SynonymsGeneration($\mathbf{T}, w_i$)
            \State $C' \gets$ FilterCandidates($\mathbf{T}, C, w_i$)
            \If{$C' \neq \emptyset$}
                \State Compute $b_k$ using Eq. (\ref{eq:bk})
                \State $N \gets N + 1$
                \If{$b_k = 1$}
                    \State $c \gets c + 1$
                \EndIf
            \EndIf
        \EndIf
    \EndFor
    \State Compute test statistic $Z$ using Eq. (\ref{eq:z})
    \State Compare $Z$ with the critical value $Z_{\alpha}$
    \State \Return watermark presence decision
\EndProcedure
\end{algorithmic}
\end{algorithm}

% \begin{algorithm}
% \small
% \caption{Fine-grained Detection Mode}\label{alg:fine_grained_detection}
% \begin{algorithmic}[1]
% \Procedure{FineGrainedDetection}{$\mathbf{T}$}
%     \State Initialize watermark pattern count $n \gets 0$
%     \State Initialize bit 1 count $c \gets 0$
%     \For{$i \in \{2, 3, \ldots, n\}$}
%         \If{POSFilter($t_i$)}
%             \State $W \gets$ SynonymsGeneration($\mathbf{T}, t_i$)
%             \State $W' \gets$ FilterCandidates($\mathbf{T}, W, t_i$)
%             \If{$W' \neq \emptyset$}
%                 \State Compute $b_k$ using Eq. (9)
%                 \State $n \gets n + 1$
%                 \If{$b_k = 1$}
%                     \State $c \gets c + 1$
%                 \EndIf
%             \EndIf
%         \EndIf
%     \EndFor
%     \State Compute test statistic $Z$ using Eq. (11)
%     \State Compare $Z$ with the critical value $Z_{\alpha}$
%     \State \Return watermark presence decision
% \EndProcedure
% \end{algorithmic}
% \end{algorithm}

% \input{src/smoothing_classification}
% \input{src/smoothing_fairness}

% \input{src/error}

\section{Experimental Evaluation} \label{sec:eval}
\begin{figure*}[t]
    \centering
    \begin{subfigure}{.25\textwidth}
        \centering
        \includegraphics[width=\linewidth]{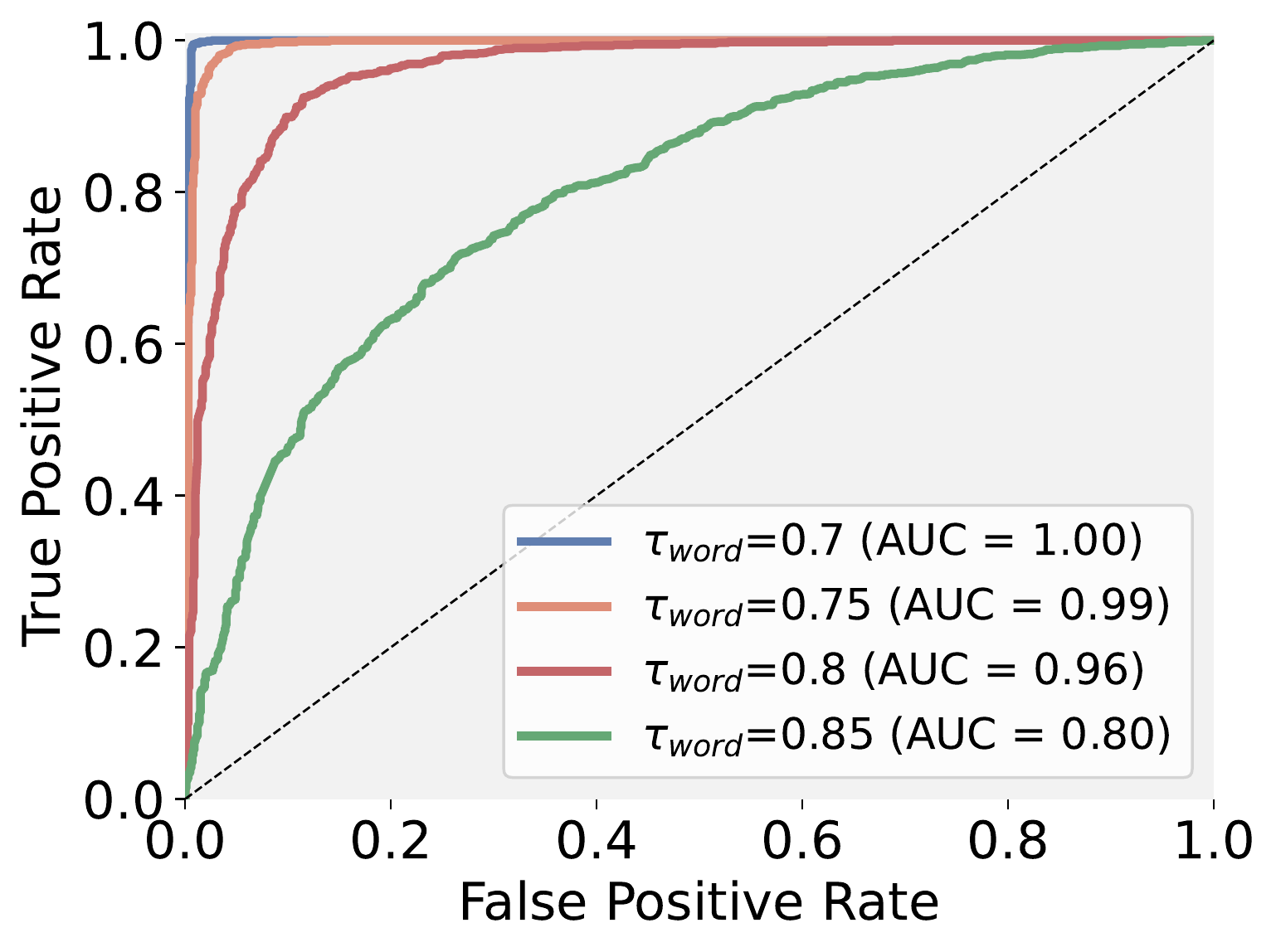}
        \caption{English \& Fast Detection}
        \label{fig:sub1}
    \end{subfigure}%
    \begin{subfigure}{.25\textwidth}
        \centering
        \includegraphics[width=\linewidth]{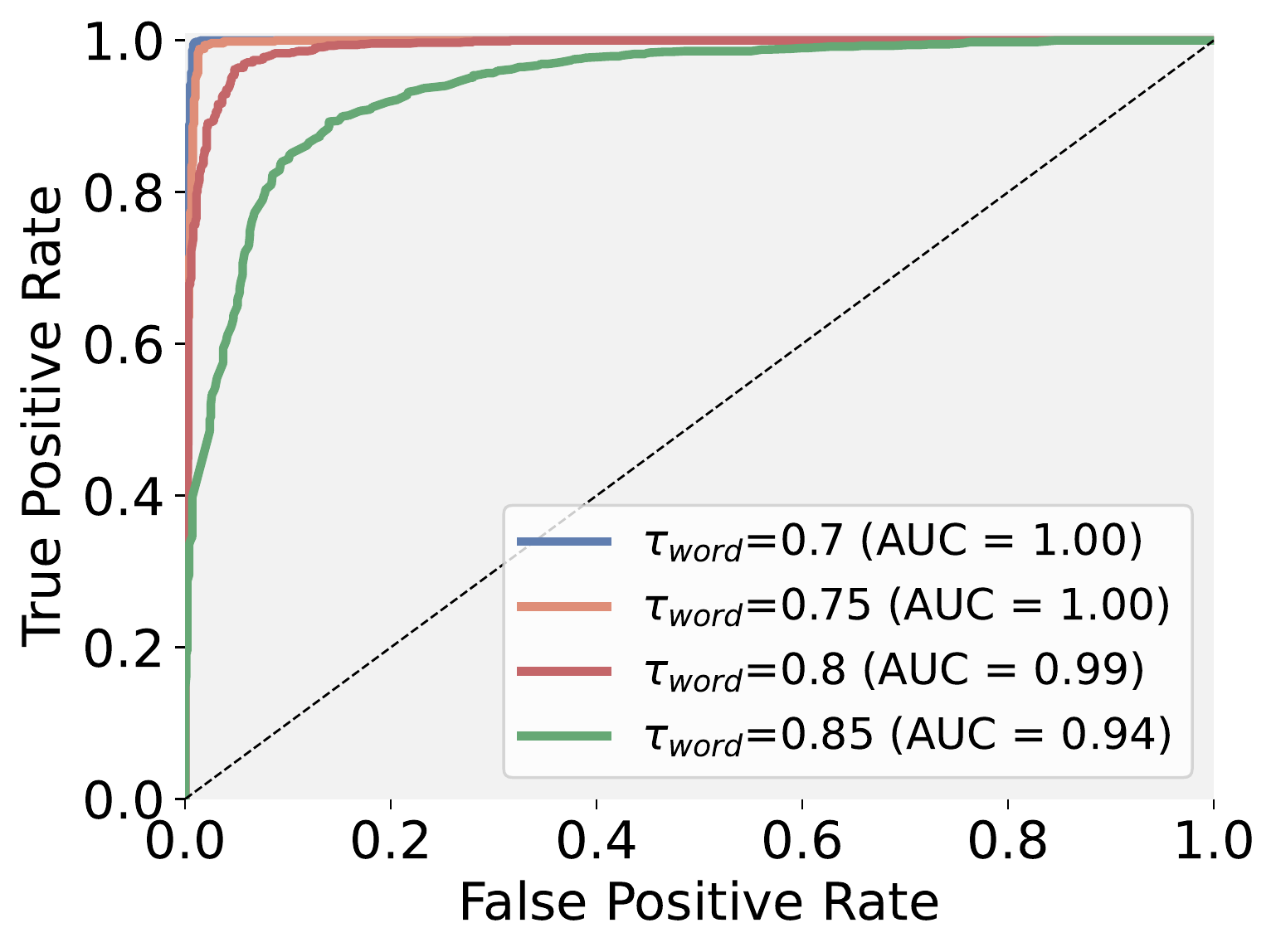}
        \caption{English \& Precise Detection}
        \label{fig:sub2}
    \end{subfigure}%
    \begin{subfigure}{.25\textwidth}
        \centering
        \includegraphics[width=\linewidth]{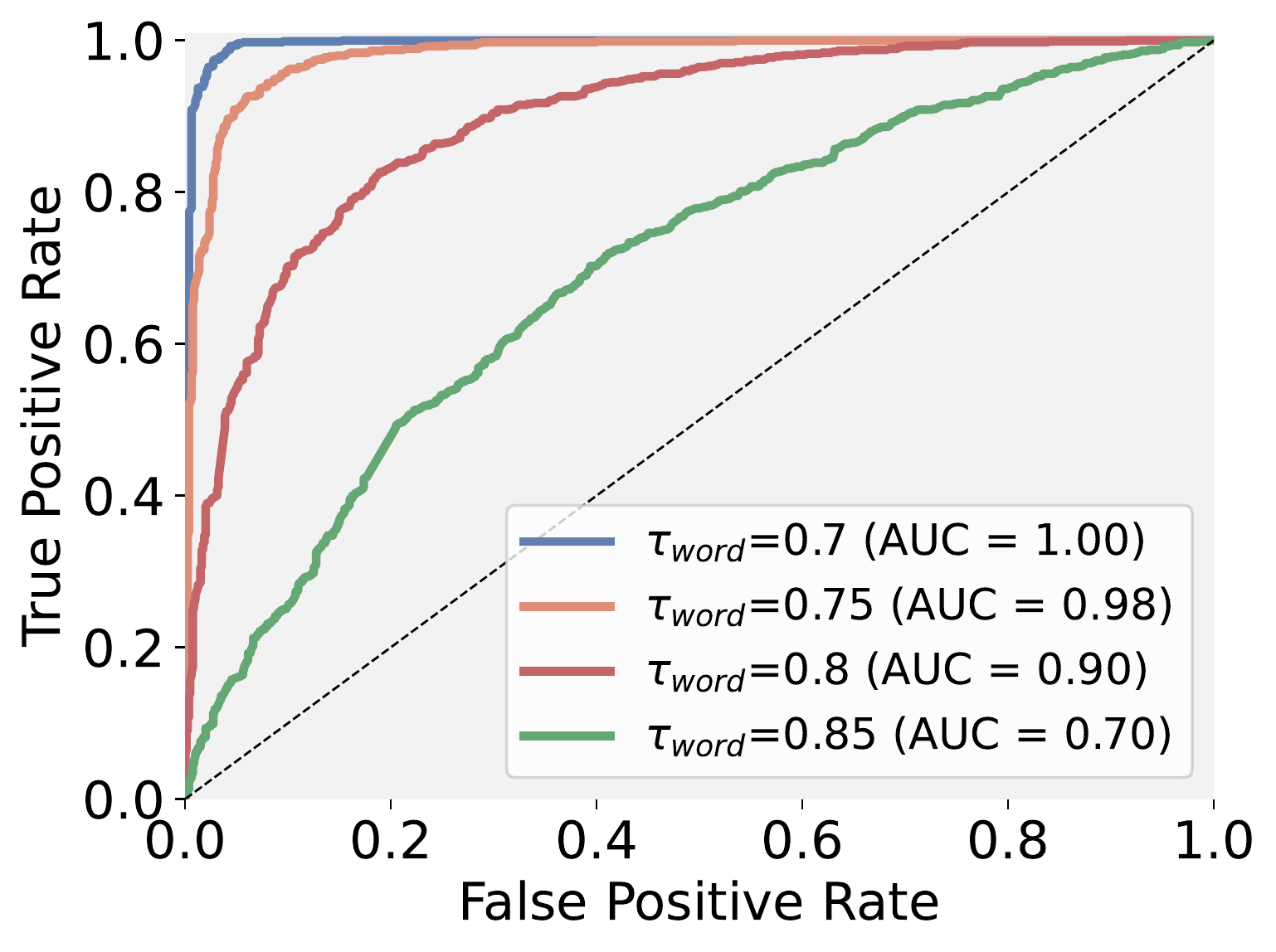}
        \caption{Chinese \& Fast Detection}
        \label{fig:sub3}
    \end{subfigure}%
    \begin{subfigure}{.25\textwidth}
        \centering
        \includegraphics[width=\linewidth]{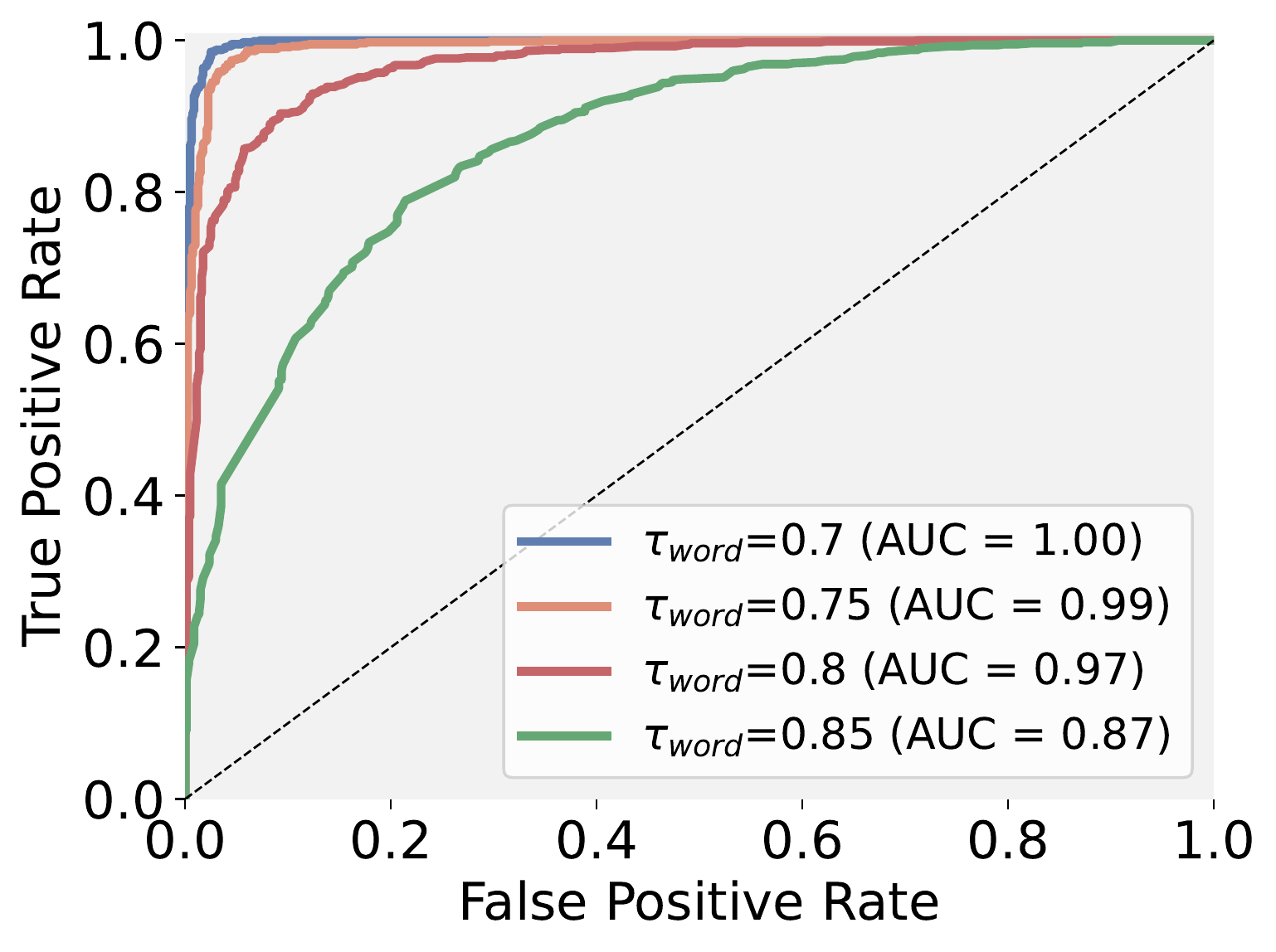}
        \caption{Chinese \& Precise Detection}
        \label{fig:sub4}
    \end{subfigure}
    \caption{ROC curves with AUC (Area Under the Curve) values for watermark detection under different languages, $\tau_{\text{word}}$ values, and detection modes. An AUC value of 1 indicates perfect classification, while a value of 0.5 implies random chance.}
    \label{fig:tau}
\end{figure*}
\subsection{Experimental Setup}
\para{Datasets}
To evaluate our method, we primarily utilize the Human ChatGPT Comparison Corpus (HC3) from \cite{hc3}. The HC3 dataset offers a crucial resource for examining linguistic and stylistic features of both human-written and ChatGPT-generated text in Chinese and English. 
We select the ChatGPT answers for evaluation. Specifically, we gather 200 samples from each of the following English subcategories: \texttt{wiki\_csai}, \texttt{open\_qa}, \texttt{medicine}, and \texttt{reddit\_eli5}. Each English sample has a length of 200±5 words. In total, we obtain 800 samples to serve as our English dataset.
Likewise, we choose 800 samples from the equivalent Chinese subcategories, including \texttt{baike}, \texttt{open\_qa}, \texttt{medicine}, and \texttt{nlpcc\_dbqa}. Each Chinese sample has a length of 200±5 characters. Note that the information-carrying capacity of 200 Chinese characters and 200 English words is different, and there are performance variations between Chinese and English language models. Thus, the results will exhibit subtle differences on these two languages. 

% Besides, treating human language systems as black-box models, we further examine the effectiveness of our method on human-written text datasets. For this purpose, we utilize the \texttt{FakeNews}\footnote{www.kaggle.com/datasets/clmentbisaillon/fake-and-real-news-dataset} dataset for English and the \texttt{People's Daily}\footnote{www.kaggle.com/datasets/concyclics/renmindaily} dataset for Chinese.
\para{Implementation Details}
We utlize the SHA-256 hashing algorithm to construct the binary encoding function.
In English experiments, we adopt the BERT model (\texttt{bert-base-cased} \cite{bert}) for synonym generation and contextualized word similarity computation. The RoBERTa model (\texttt{roberta-large-mnli} \cite{roberta}) is employed for sentence similarity calculation, while the Word-to-Vec model (\texttt{glove-wiki-gigaword-100} \cite{glove}) is used for global word similarity assessment.
Similarly, in Chinese experiments, we employ the word-level Chinese BERT model (\texttt{wobert\_chinese\_plus\_base} \cite{wobert}), the Chinese RoBERTa model (\texttt{Erlangshen-Roberta-330M-Similarity} \cite{erlangshen}), and the Chinese Word-to-Vec model (\texttt{sgns.merge.word}\cite{cnembeddings}).

Regarding the hyperparameters, we set $\lambda=0.83$ and $K=32$ as default.
The sentence similarity $S_{\text{sent}}$ between the watermarked text and the original text typically remains stable, as they differ by merely a few words. The $S_{\text{sent}}$ score will decrease substantially when antonyms are involved. Therefore, in our primary experiments, we fix $\tau_{\text{sent}}=0.8$ and focus on investigating the impact of varying $\tau_{\text{word}}$ values. Moreover, we conduct an ablation study to 
% demonstrate the 
% \yx{importance of sentence-level and word-level semantic similarity constraints.}
assess the roles of $\tau_\text{sent}$ and $\tau_\text{word}$ in semantic quality control.

\para{Metrics}
To demonstrate detection performance, we primarily employ Receiver Operating Characteristic (ROC) curves to present detection results. ROC curves are graphical plots that showcase the diagnostic ability of a binary classifier as its discrimination threshold varies, illustrating the true positive rate against the false positive rate and offering insights into the sensitivity-specificity trade-off. Besides, to evaluate robustness, we use $Z$-score to represent watermark strength.
For fidelity assessment, we employ the METEOR score \cite{meteor}, a traditional metric widely used in machine translation to compare output sentences with references which conducts n-gram alignments between reference and output text. Scores range from 0 to 1 (identical sentences). However, METEOR alone is insufficient for evaluating semantics, as two sentences with an equal number of changed words may have similar METEOR scores but differ in semantics. Thus, we employ the language models (\ie \texttt{all-MiniLM-L6-v2}\footnote{\url{https://huggingface.co/sentence-transformers/all-MiniLM-L6-v2}} for English and \texttt{Erlangshen-Roberta-330M-Si\\milarity}\footnote{\url{https://huggingface.co/IDEA-CCNL/Erlangshen-Roberta-330M-Similarity}} for Chinese) to approximate semantic similarity between original and watermarked text.

\subsection{Watermark Strength under Different $\tau_{\text{word}}$} \label{ssec:Watermark Strength}
To investigate the influence of different $\tau_{\text{word}}$ values on watermark strength, we perform watermark injection and detection on Chinese and English samples from HC3 dataset using different $\tau_{\text{word}}$ values, specifically 0.7, 0.75, 0.8, and 0.85. Under such settings, the ROC curves in Figure \ref{fig:tau} depict the experimental results. As can be observed, for identical language and detection mode, higher $\tau_{\text{word}}$ values result in weaker watermark strength (\ie smaller AUC values). This is because higher $\tau_{\text{word}}$ values enforce stricter constraints on the generated synonyms, resulting in fewer modifiable words and weakened watermark strength. On the other hand, lower $\tau_{\text{word}}$ values permit more relaxed synonym constraints, yielding more modifiable words, stronger watermarks, but may compromise the original semantic quality, which will be discussed in \cref{ssec:quality}.
Moreover, with identical language and $\tau_{\text{word}}$ value, the precise detection surpasses the fast detection, indicating that conducting further analysis on words can effectively enhance the detection capabilities.

\subsection{Fidelity Analysis}
\begin{figure}[t]
    \centering
    \includegraphics[width=0.8\linewidth]{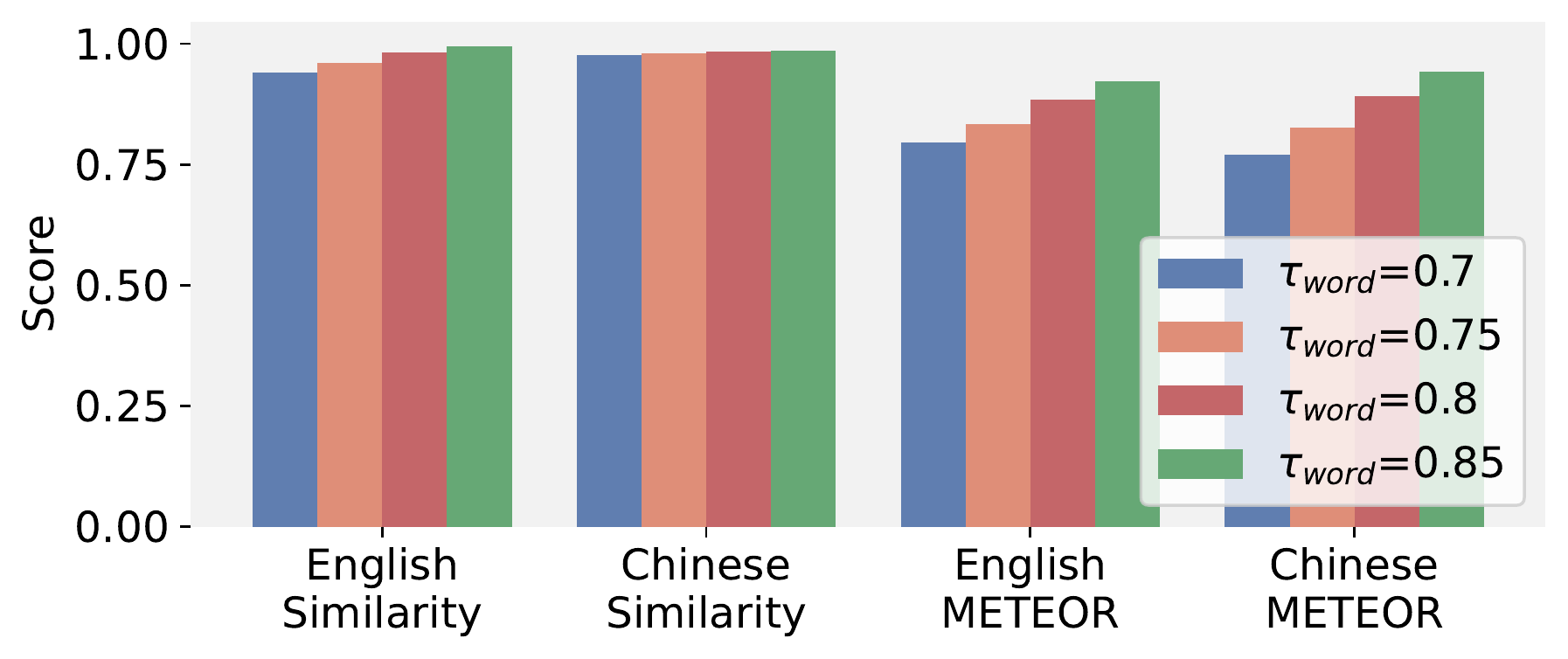}
    \caption{Semantic similarity and METEOR scores of watermarked text compared to the original text under different $\tau_{\text{word}}$ values.}
    \label{fig:quality}
\end{figure}
\begin{figure}[t]
    \centering
    \begin{subfigure}[b]{0.48\linewidth}
        \centering
        \includegraphics[width=\linewidth]{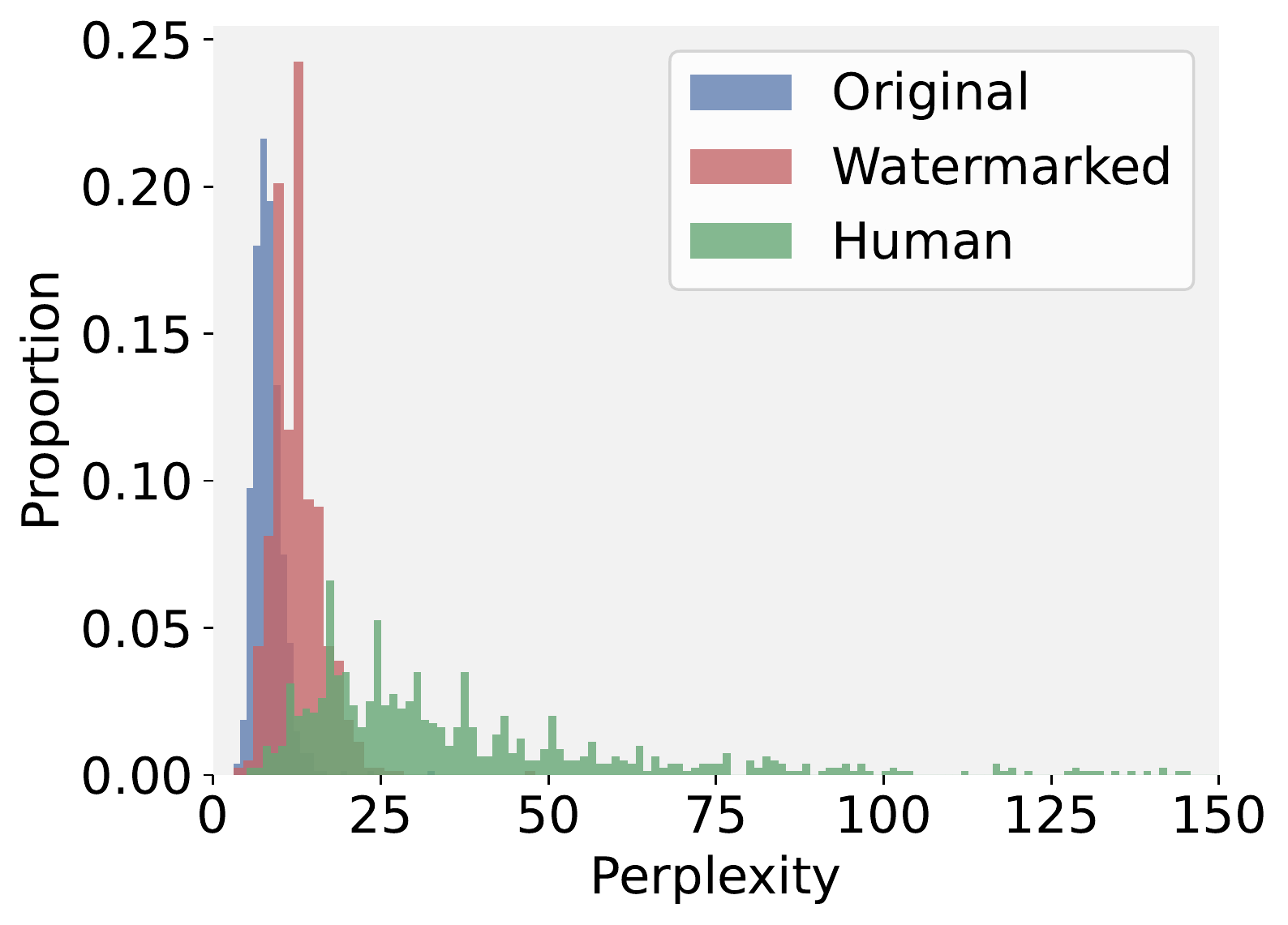}
        \caption{English}
    \end{subfigure}
    % \hspace{0.05\linewidth} % Adjust the horizontal space between the subfigures, if necessary
    \begin{subfigure}[b]{0.48\linewidth}
        \centering
        \includegraphics[width=\linewidth]{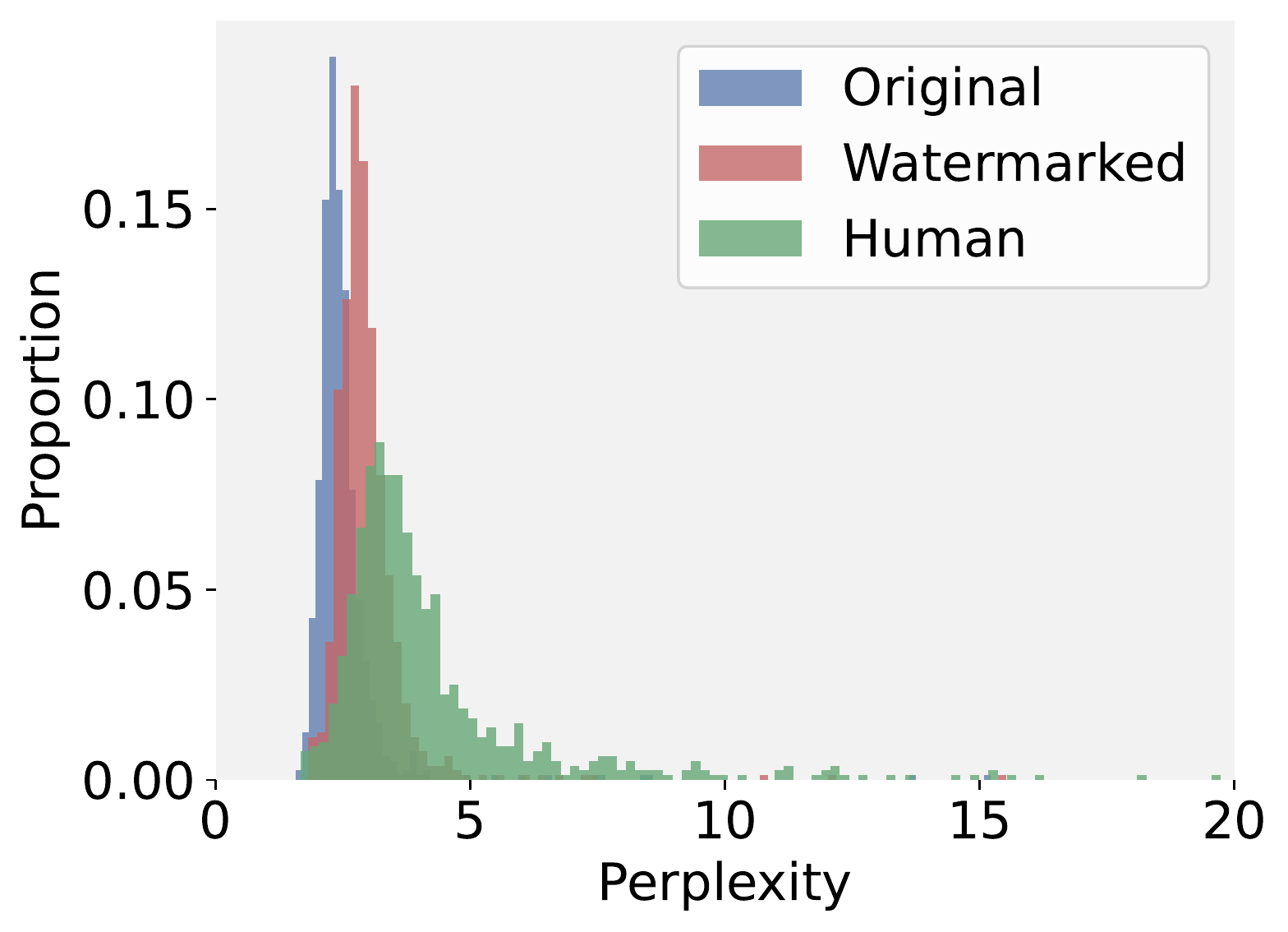}
        \caption{Chinese}
    \end{subfigure}
    \caption{Perplexity distributions of original generated text, watermarked generated text, and human-written text.}
    \label{fig:ppl}
\end{figure}

\para{Semantic Quality} \label{ssec:quality}
We present the semantic similarity scores and METEOR scores for watermarked text in comparison to the original text under various $\tau_{\text{word}}$ values in Figure \ref{fig:quality}. Higher $\tau_{\text{word}}$ values result in fewer alterations to the original text, making the watermarked text closer to the original while maintaining high semantic quality. Furthermore, a decrease in $\tau_{\text{word}}$ value does not lead to a significant reduction in semantic similarity because language models focus more on the overall semantic similarity of the text, whereas METEOR scores decline more rapidly as they are highly sensitive to word substitutions. Taking both watermark strength and semantic quality into consideration, we choose $\tau_{\text{word}}=0.8$ for English and $\tau_{\text{word}}=0.75$ for Chinese as default values for subsequent experiments.
\para{Perplexity Distributions}
Perplexity (PPL) is a widely used metric for evaluating language model performance, defined as the exponential of the negative average log-likelihood of a given text under the language model. Lower PPL values indicate a more confident language model. Language models are trained on extensive text corpora, enabling them to learn common language patterns and structures. Thus, PPL can be used to assess how well a text conforms to typical characteristics.
We employ Chinese and English GPT-2 models (\texttt{Wenzhong-GPT2-110M}\footnote{\url{https://huggingface.co/IDEA-CCNL/Wenzhong-GPT2-110M}} for Chinese and \texttt{gpt2-medium}\footnote{\url{https://huggingface.co/gpt2-medium}} for English) to calculate the \yx{perplexity distributions of original generated text, watermarked generated text, and human-written text. This allows us to examine the changes introduced by watermark injection from the perspective of perplexity. Figure \ref{fig:ppl} shows that original generated text exhibits lower PPL values compared to human-written text, as language models excel at reproducing common patterns and structures. In contrast, humans can express themselves in diverse ways, challenging GPT-2's prediction capabilities. Therefore, human-written text exhibit higher PPL values and display a long-tailed distribution.
Moreover, due to the increased lexical diversity resulting from our watermark injection, the PPL distribution of watermarked generated text falls between original generated text and human-written text.}

\begin{figure}[t]
    \centering
    \begin{subfigure}[b]{0.45\linewidth}
        \centering
        \includegraphics[width=\linewidth]{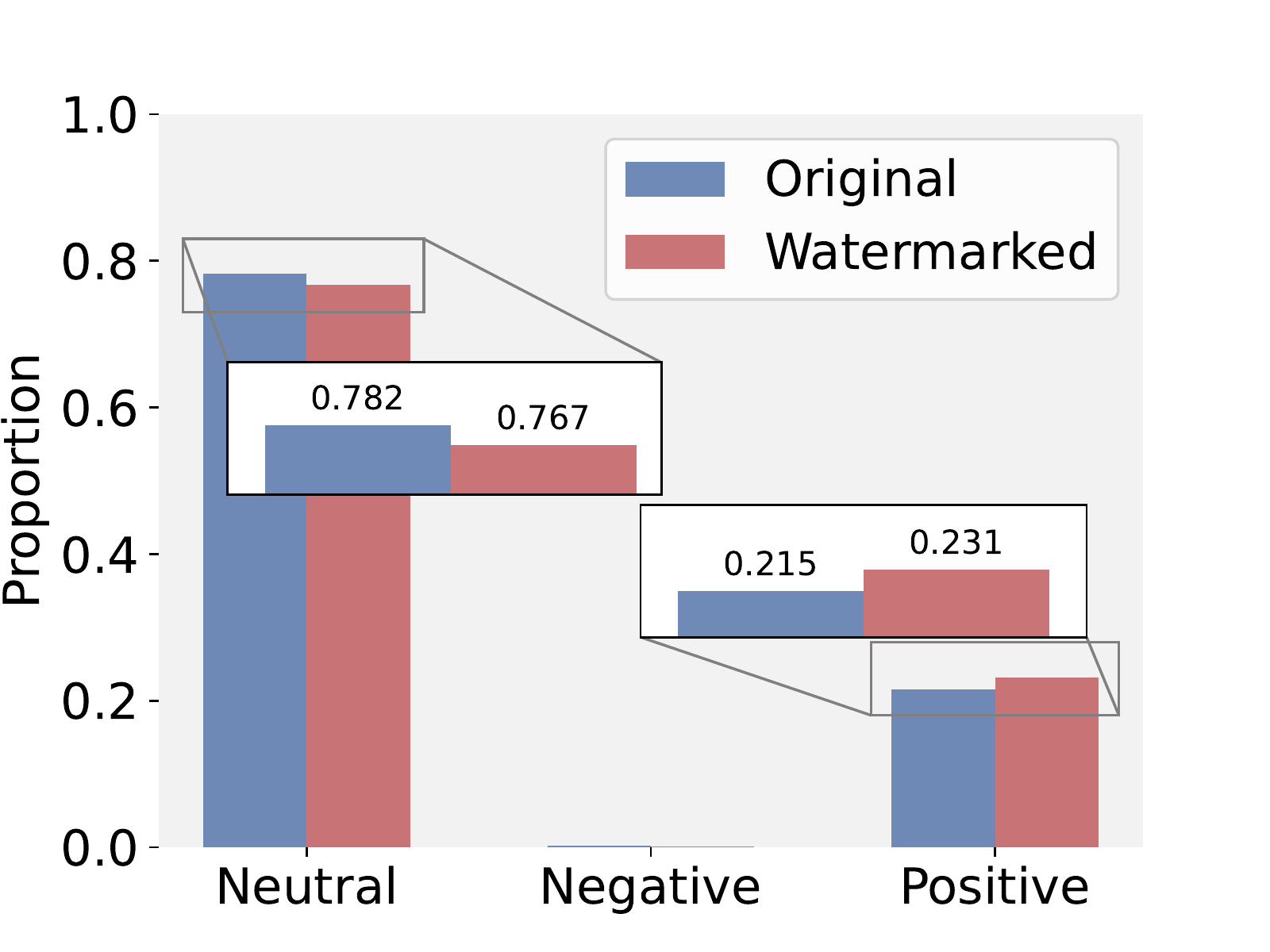}
        \caption{English}
    \end{subfigure}
    % \hspace{0.05\linewidth} % Adjust the horizontal space between the subfigures, if necessary
    \begin{subfigure}[b]{0.45\linewidth}
        \centering
        \includegraphics[width=\linewidth]{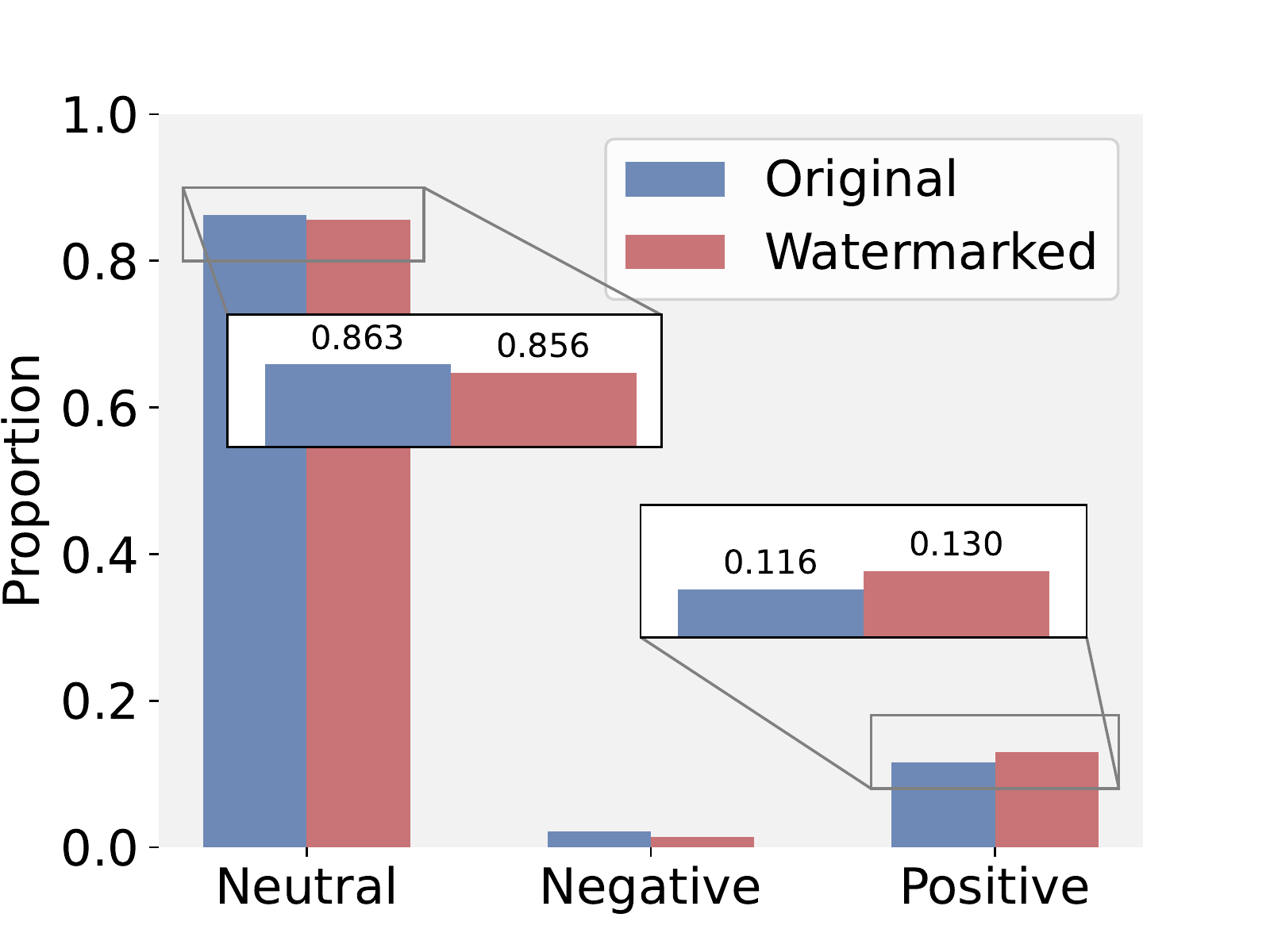}
        \caption{Chinese}
    \end{subfigure}
    \caption{Sentiment distributions of original and watermarked text.}
    \label{fig:senti}
\end{figure}

\para{Sentiment Distributions}
We visualize the sentiment of watermarked and original texts to illustrate that watermark injection has minimal impact on the original sentiment distribution. We conduct sentiment analysis on both English and Chinese datasets using a multilingual sentiment classification model (\texttt{twitter-xlm-roberta-\\base-sentiment}\footnote{\url{https://huggingface.co/cardiffnlp/twitter-xlm-roberta-base-sentiment}}) fine-tuned on a Twitter corpus.
By comparing the sentiment distributions of watermarked and original text in Figure \ref{fig:senti}, we observe that the impact of watermark injection on the overall sentiment is negligible. This finding indicates that our method effectively preserves the original sentiment of the text while injecting the watermark, ensuring that the watermarked text remains true to the intended emotion.
\begin{figure*}[t]
    \centering
    \includegraphics[width=\linewidth]{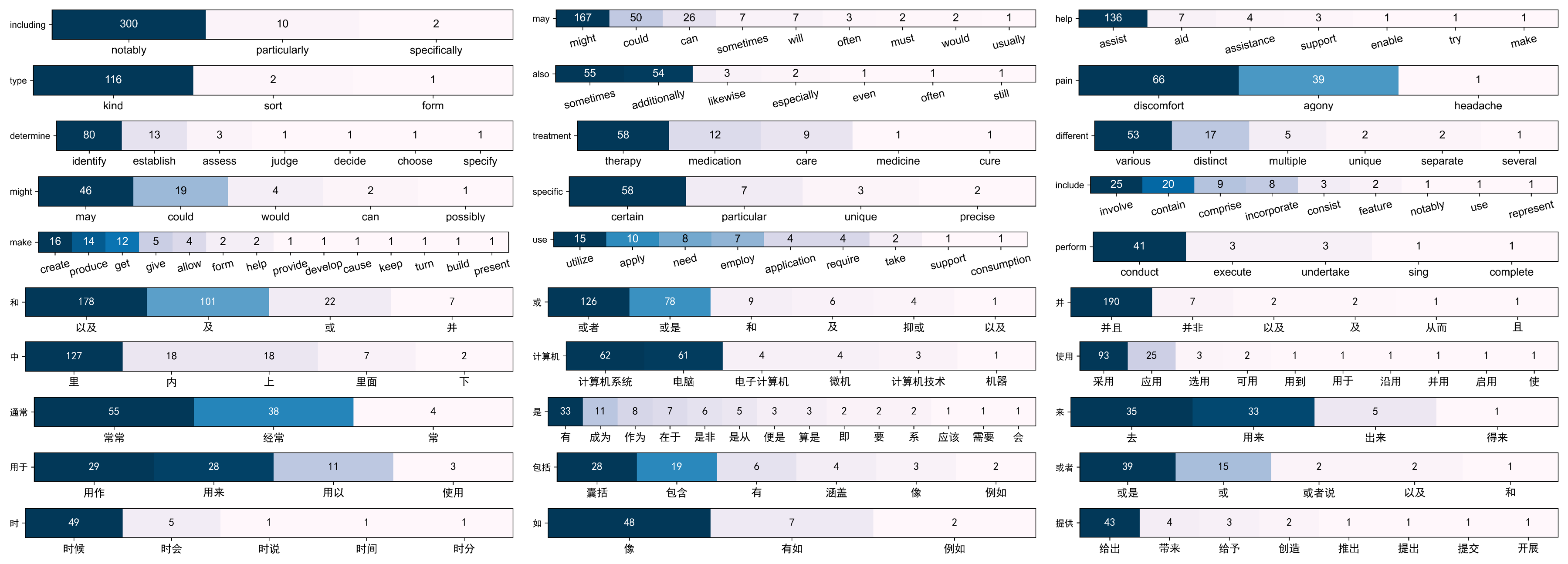}
    \caption{Replacement frequency heatmap. Each row begins with an original word, and each cell indicates the frequency of its corresponding replacement for watermark injection. The color intensity represents the frequency, with darker shades indicating higher frequencies.}
    \label{fig:matrix}
\end{figure*}

\subsection{Qualitative Analysis}
\para{Exploring Word Substitutions in Watermark Injection}
To gain deeper insights into the modifications resulting from watermark injection, we visualize the subsituted words alongside their corresponding substitutions. As shown in Figure \ref{fig:matrix}, we display the most frequently substituted words and their respective substitutions for both English and Chinese. For more complete examples of original and watermarked texts, please refer to the additional materials. The following observations can be drawn from the visualizations:1) Watermark injection does not follow a fixed substitution rule; instead, it dynamically changes based on the context, meaning a word is not consistently replaced by a specific substitute.
2) The substitutions exhibit similarities between English and Chinese but subtle distinctions arise due to the unique linguistic properties of each language. For example, English substitutions primarily consist of verbs, adjectives, and adverbs with distinct roots but similar meanings, whereas Chinese substitutions display greater flexibility, encompassing not only verbs, adjectives, adverbs, and nouns but also conjunctions with multiple equivalent alternatives (\eg ``\begin{CJK*}{UTF8}{gbsn}\textbf{或}\end{CJK*}''-``\begin{CJK*}{UTF8}{gbsn}或者\end{CJK*}'' and ``\begin{CJK*}{UTF8}{gbsn}如果\end{CJK*}''-``\begin{CJK*}{UTF8}{gbsn}假如\end{CJK*}'').

% \para{Word-frequency Distributions}
\subsection{Watermark Strength under Different Text Lengths}
\begin{figure}[t]
    \centering
    \begin{subfigure}[b]{0.48\linewidth}
        \centering
        \includegraphics[width=\linewidth]{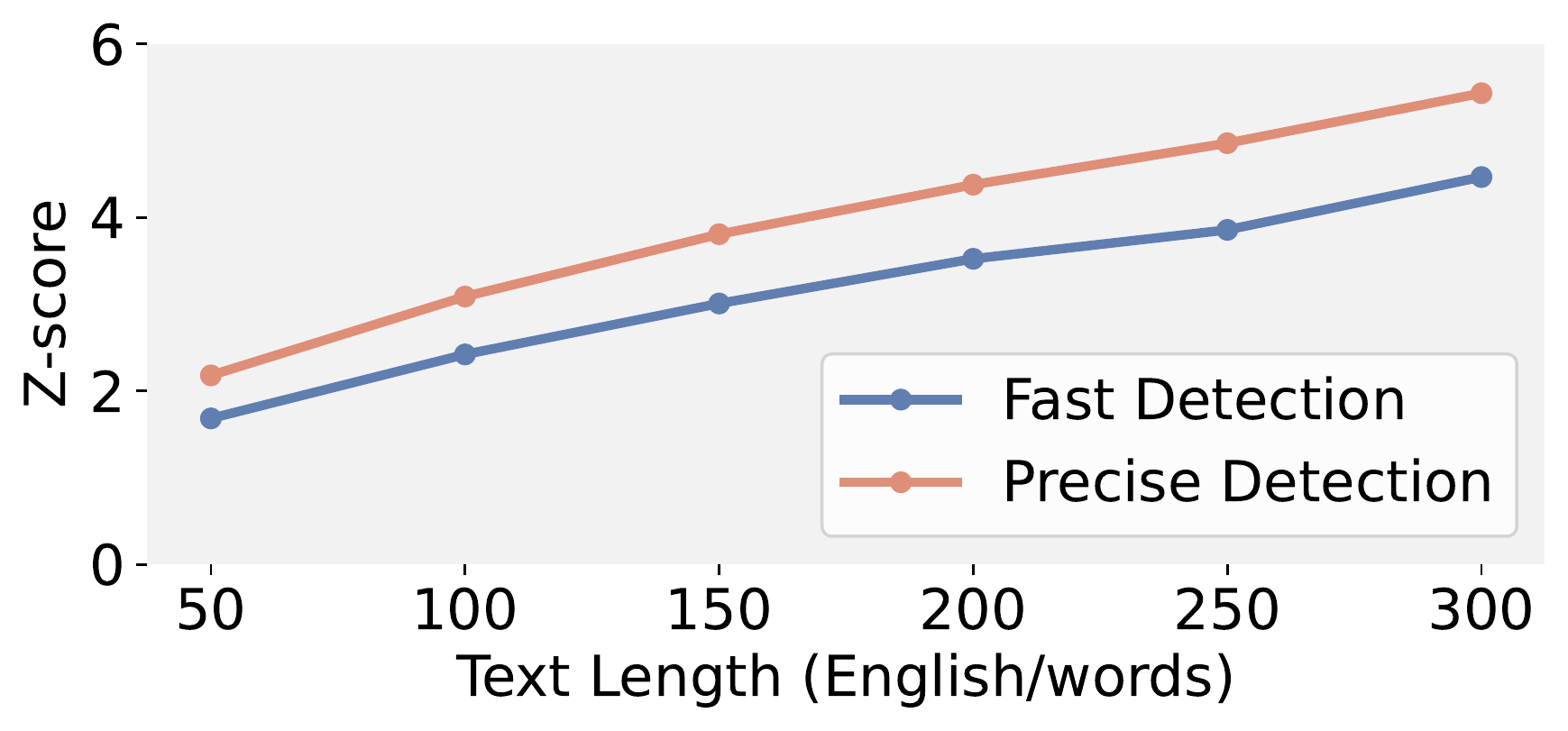}
        \caption{English}
    \end{subfigure}
    % \hspace{0.05\linewidth} % Adjust the horizontal space between the subfigures, if necessary
    \begin{subfigure}[b]{0.48\linewidth}
        \centering
        \includegraphics[width=\linewidth]{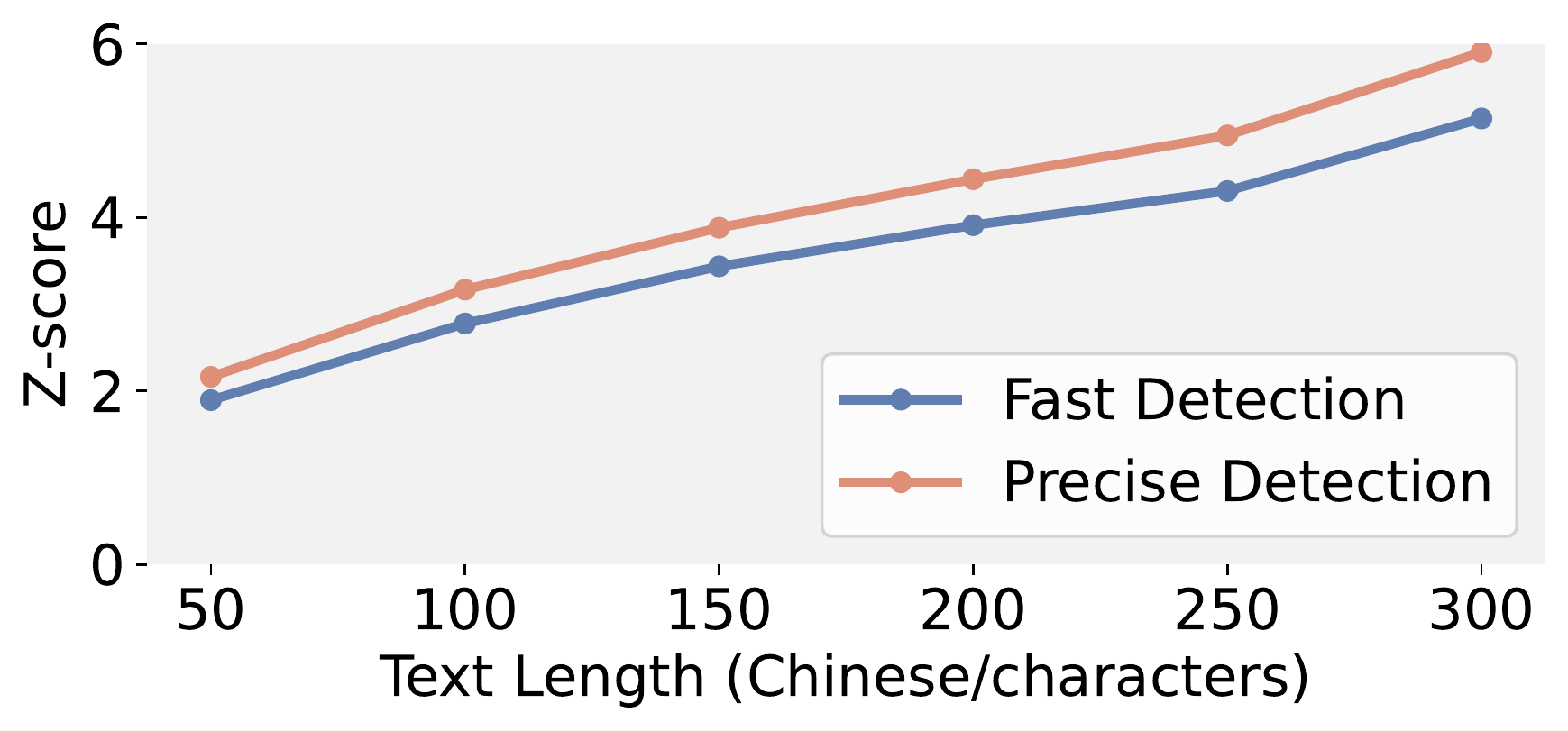}
        \caption{Chinese}
    \end{subfigure}
    \caption{Watermark strength under different text lengths.}
    \label{fig:length}
\end{figure}

We believe that the watermark strength tends to increase as the text length grows, providing more words available for watermark injection. To verify this, we select samples of varying lengths (ranging from 50 words to 300 words) in the English dataset, with 800 samples for each length, to perform watermark injection and detection. Similarly, we select samples of different lengths (ranging from 50 characters to 300 characters) in the Chinese dataset. We employ $Z$-score to measure watermark strength. As shown in Figure \ref{fig:length}, for both English and Chinese, there is a clear trend showing that as the text length increases, the watermark strength, represented by the $Z$-score, also increases. This observation supports that longer texts offer more opportunities for watermark injection.
When comparing the results of fast detection and precise detection, it is evident that the precise mode consistently yields higher $Z$-scores across all text lengths for both languages. Moreover, as the text length increases, the enhancement in detection performance becomes more significant compared to shorter texts.

\begin{figure}[t]
    \centering
    \begin{subfigure}{.25\textwidth}
        \centering
        \includegraphics[width=\linewidth]{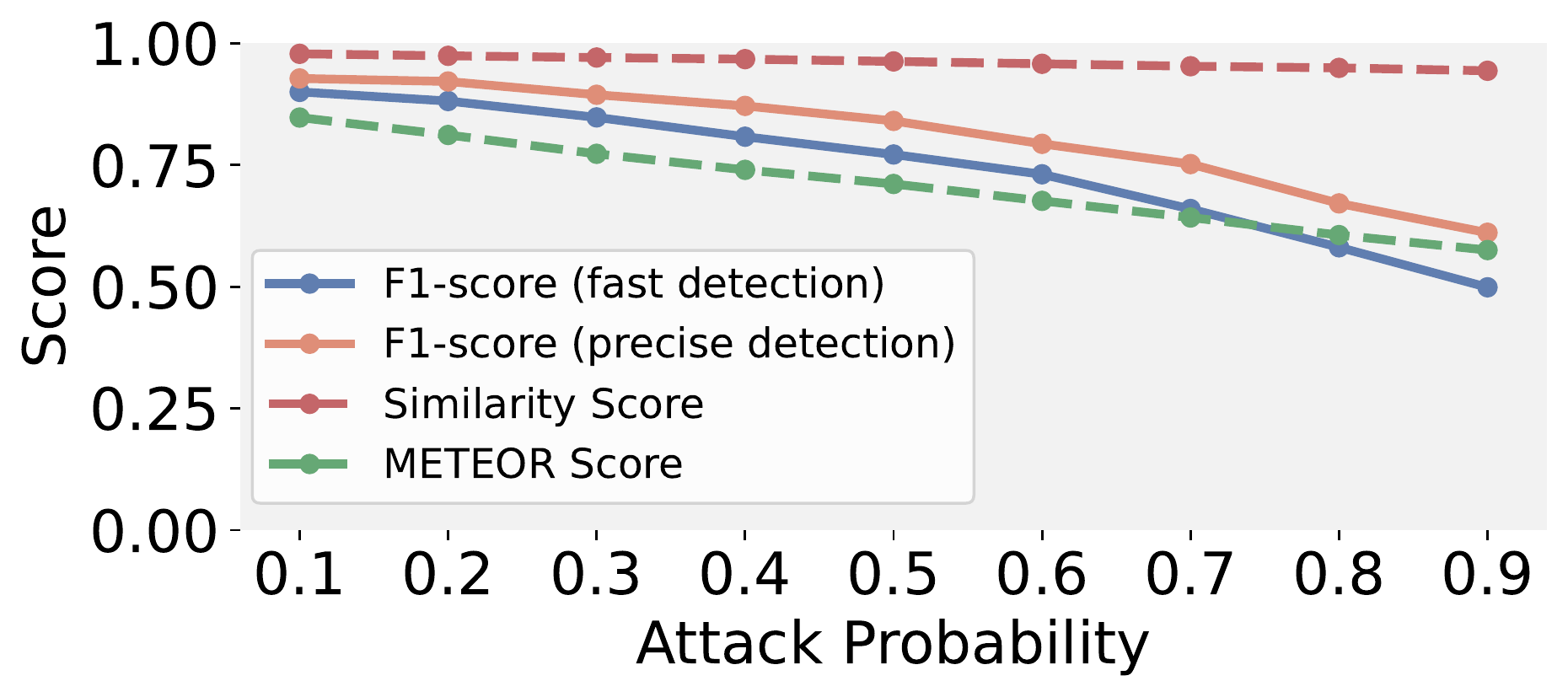}
        \caption{English \& Baidu Translator}
        \label{fig:sub1}
    \end{subfigure}%
    \begin{subfigure}{.25\textwidth}
        \centering
        \includegraphics[width=\linewidth]{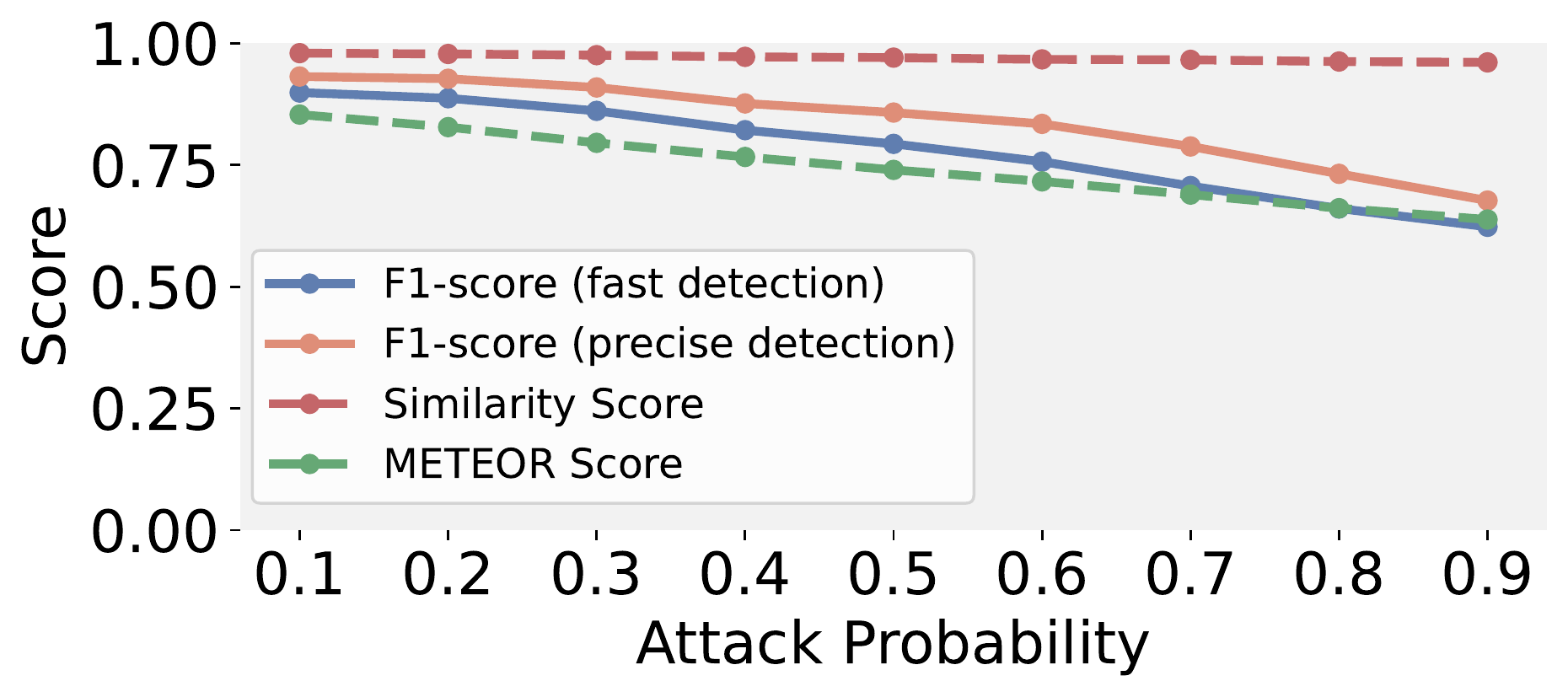}
        \caption{English \& DeepL Translator}
        \label{fig:sub2}
    \end{subfigure}%
    
    \begin{subfigure}{.25\textwidth}
        \centering
        \includegraphics[width=\linewidth]{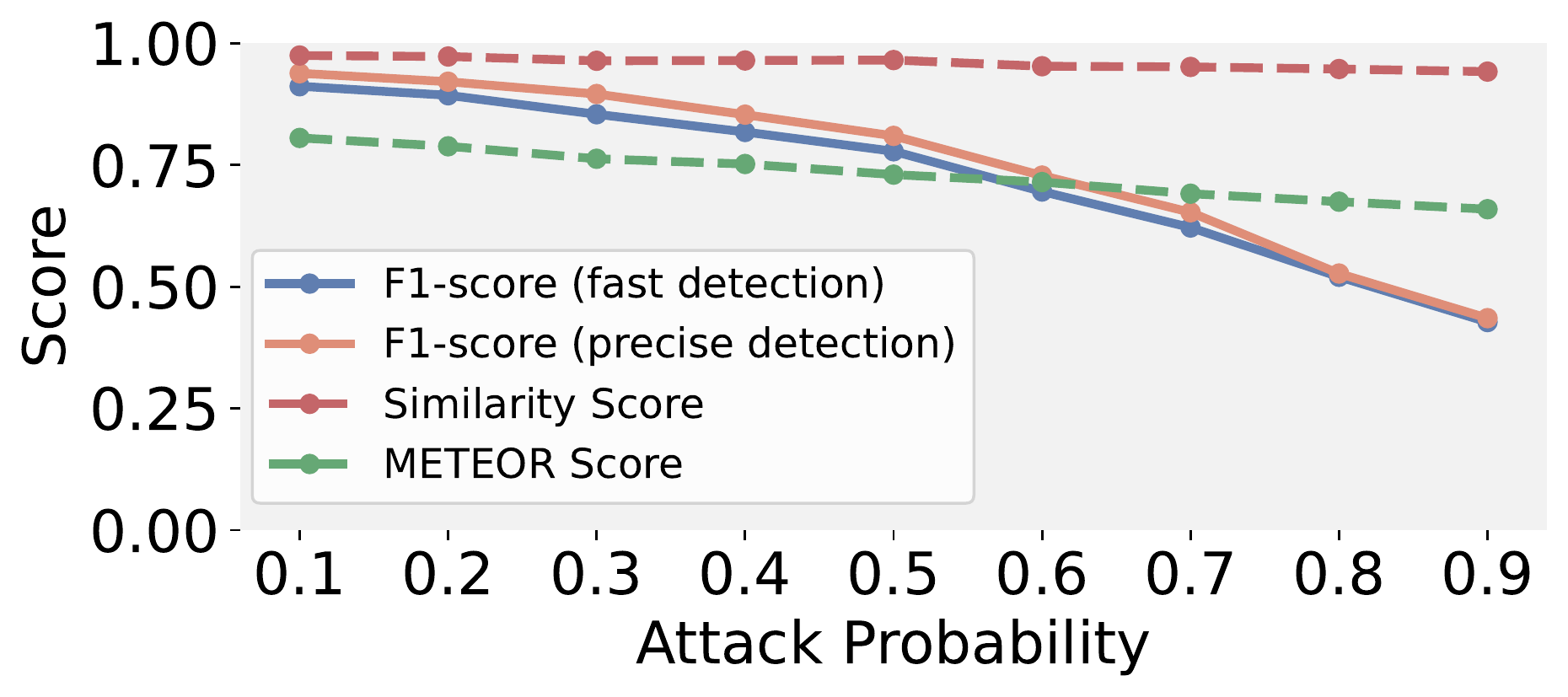}
        \caption{Chinese \& Baidu Translator}
        \label{fig:sub3}
    \end{subfigure}%
    \begin{subfigure}{.25\textwidth}
        \centering
        \includegraphics[width=\linewidth]{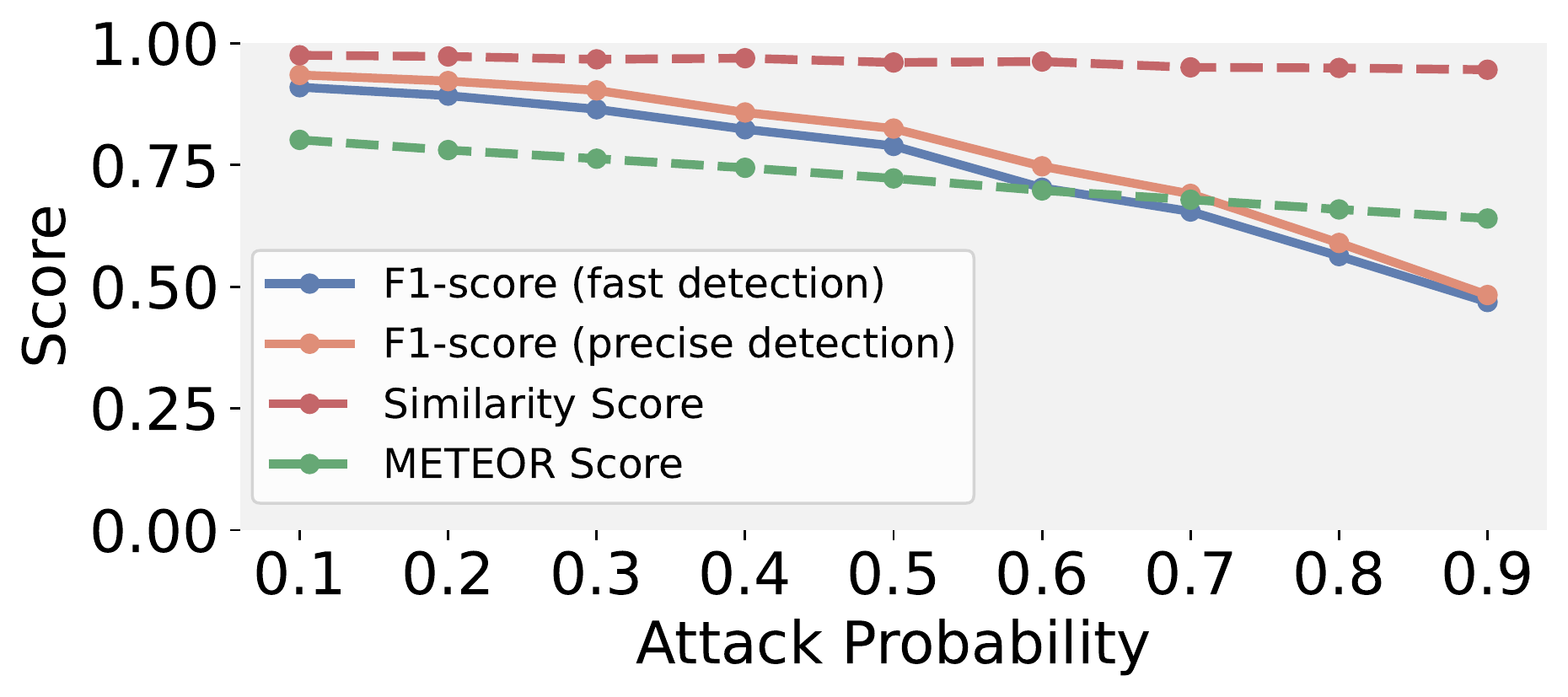}
        \caption{Chinese \& DeepL Translator}
        \label{fig:sub4}
    \end{subfigure}
    \caption{Robustness analysis of the watermark under re-translation attacks. The x-axis represents the attack probability. The y-axis displays the F1-score, similarity score, and METEOR score for both fast and precise detection modes. Higher scores indicate better performance.}
    \label{fig:re_trans_attack}
\end{figure}
\subsection{Robustness Analysis}
We simulate two primary types of attacks that an attacker might use to remove the watermark, specifically sentence-level attacks (including re-translation and polishing) and  word-level attacks (including word deletion and synonym substitution). To illustrate the impact of attacking varying proportions of text content, we set the probability of each sentence or word being subjected to an attack. Following this, we assess the watermark strength and the semantic quality of the watermarked text under different attack probabilities.
\para{Re-translation Attack}
Considering a scenario in which attackers attempt to re-translate portions of the watermarked text, we calculate the F1-score (with the significance level $\alpha=0.05$), semantic similarity score, and METEOR score of the watermarked text under different attack probabilities. We employ two commercial translation tools, namely Baidu Translator\footnote{\url{https://fanyi.baidu.com/}} and DeepL Translator\footnote{\url{https://www.deepl.com/translator}} to perform the attack. For English text, we first translate it to Chinese and then translate the resulting Chinese text back to English. Conversely, for Chinese text, we first translate it to English and then translate the resulting English text back to Chinese. As shown in Figure \ref{fig:re_trans_attack}, as the attack probability increases, more content within the text is altered, resulting in a progressively weakened watermark. However, the METEOR scores follow a similar trend. This suggests that when a large proportion of the text is modified, the watermark will be weakened, but the attacked text also undergoes significant structural and semantic changes compared to the original text. This contradicts the attacker's objective. Consequently, when the attacker makes small-scale attacks on the text content, our watermark demonstrates good robustness. For instance, when half of the text undergoes a re-translation attack, the F1-score remains above 0.75, still maintaining a reasonable level of robustness.
\begin{figure}[t]
    \centering
    \begin{subfigure}[b]{0.8\linewidth}
        \centering
        \includegraphics[width=\linewidth]{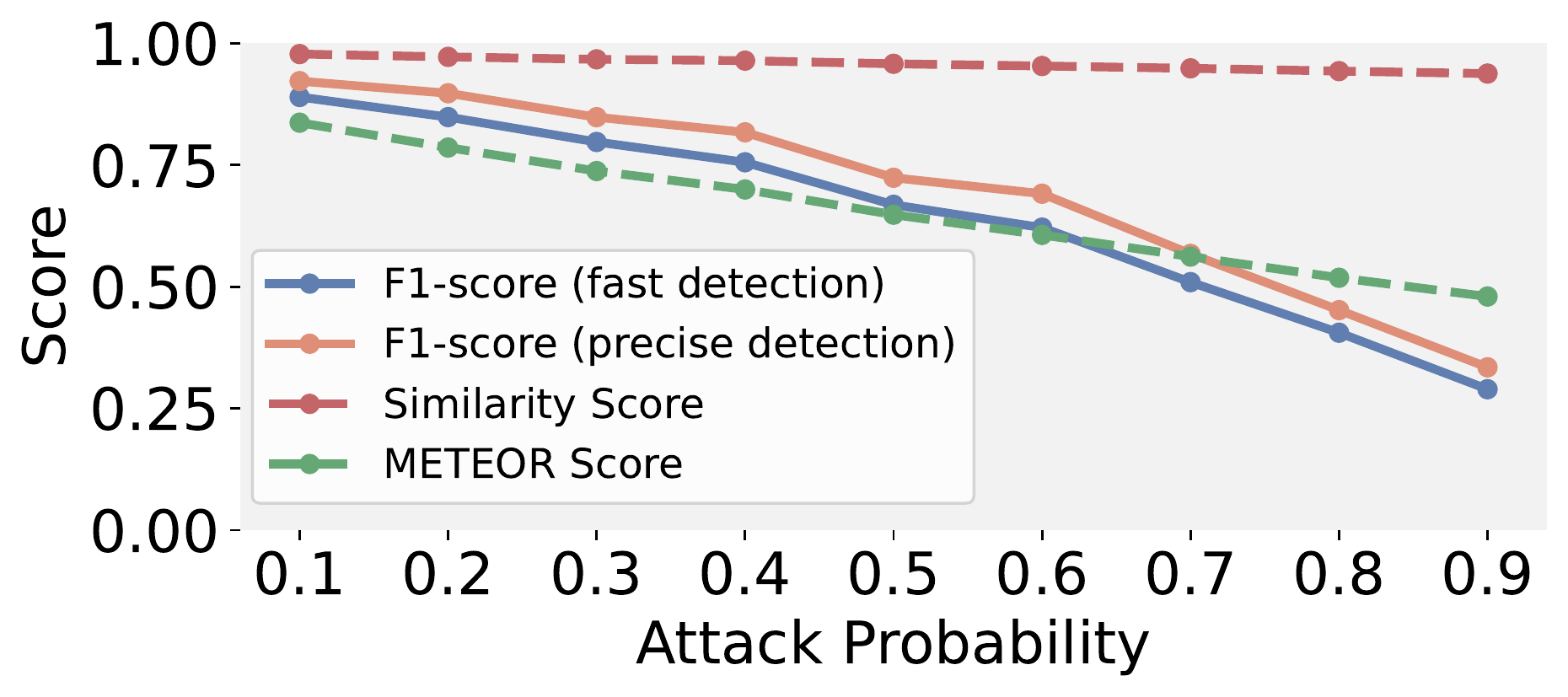}
        \caption{English}
    \end{subfigure}
    % \hspace{0.05\linewidth} % Adjust the horizontal space between the subfigures, if necessary
    \begin{subfigure}[b]{0.8\linewidth}
        \centering
        \includegraphics[width=\linewidth]{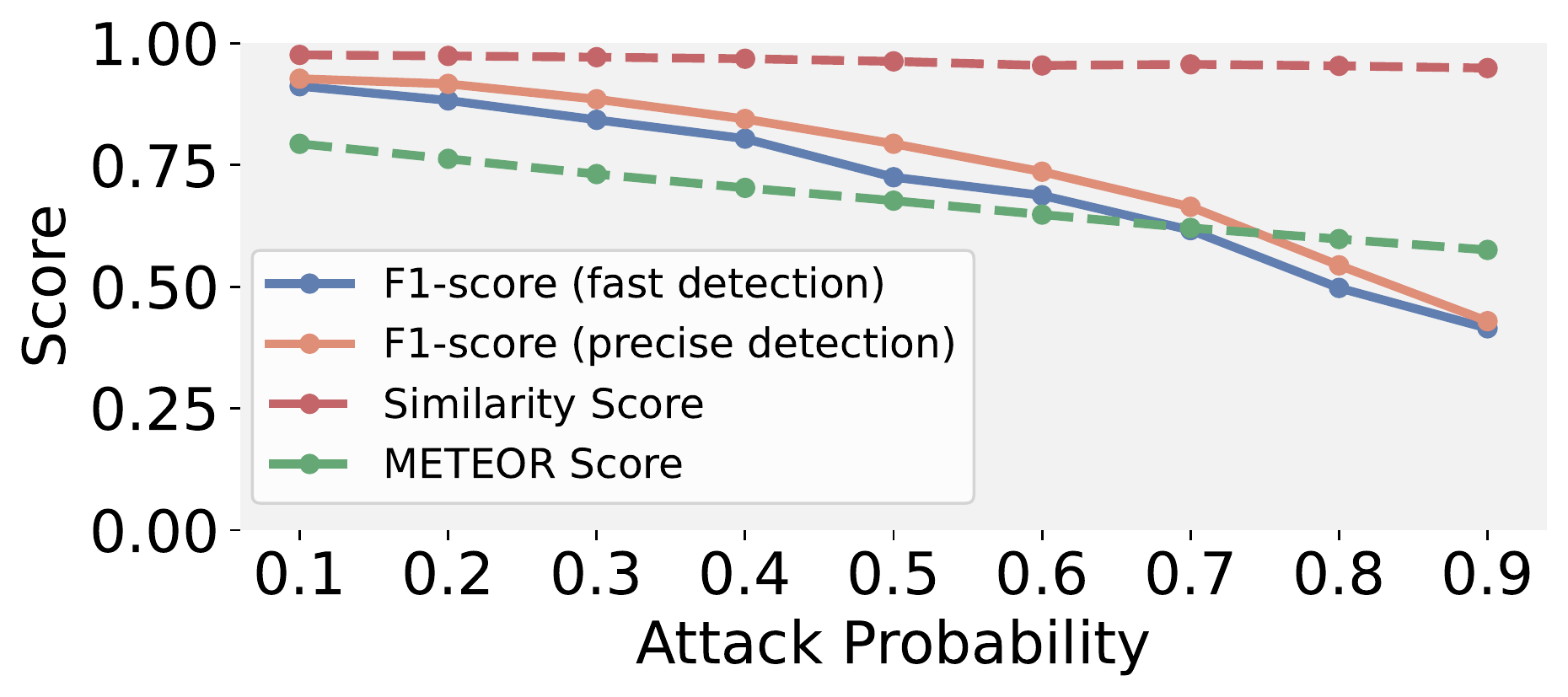}
        \caption{Chinese}
    \end{subfigure}
    \caption{Robustness analysis of the watermark under polishing attacks.}
    \label{fig:polish_attack}
\end{figure}
\para{Polishing Attack}
Polishing attacks on watermarked texts can be carried out by attackers using top-notch LLM services. We employ the API of \texttt{GPT-3.5-turbo} provided by OpenAI to perform this type of attack. We calculate the F1-score (with $\alpha=0.05$), semantic similarity score, and METEOR score of the watermarked text under different attack probabilities. The prompts we use for English and Chinese text are: ``Please polish the input text without changing its meaning. The input text is: [watermarked text]'' and ``\begin{CJK*}{UTF8}{gbsn}润色这段话，不要改变原始语义。这段话的内容是: \end{CJK*}[watermarked text]'', respectively. The results, shown in Figure \ref{fig:polish_attack}, demonstrate that GPT-3.5-based polishing causes more severe damage to the watermark than re-translation. This is mainly because polishing alters the text content to a greater extent than translation, and the GPT-3.5 model may generate more associative content based on the original content. 
% Besides, it sometimes refuses to answer or respond in English to Chinese, introducing more noise. 
However, this also leads to more significant damage to the semantic quality of the original text, as indicated by the METEOR score drop to around 0.5 when attacked with a probability of 0.6, which indicates that the alignment between the original text and the attacked text is significantly affected.

\begin{figure}[t]
    \centering
    \begin{subfigure}[b]{0.8\linewidth}
        \centering
        \includegraphics[width=\linewidth]{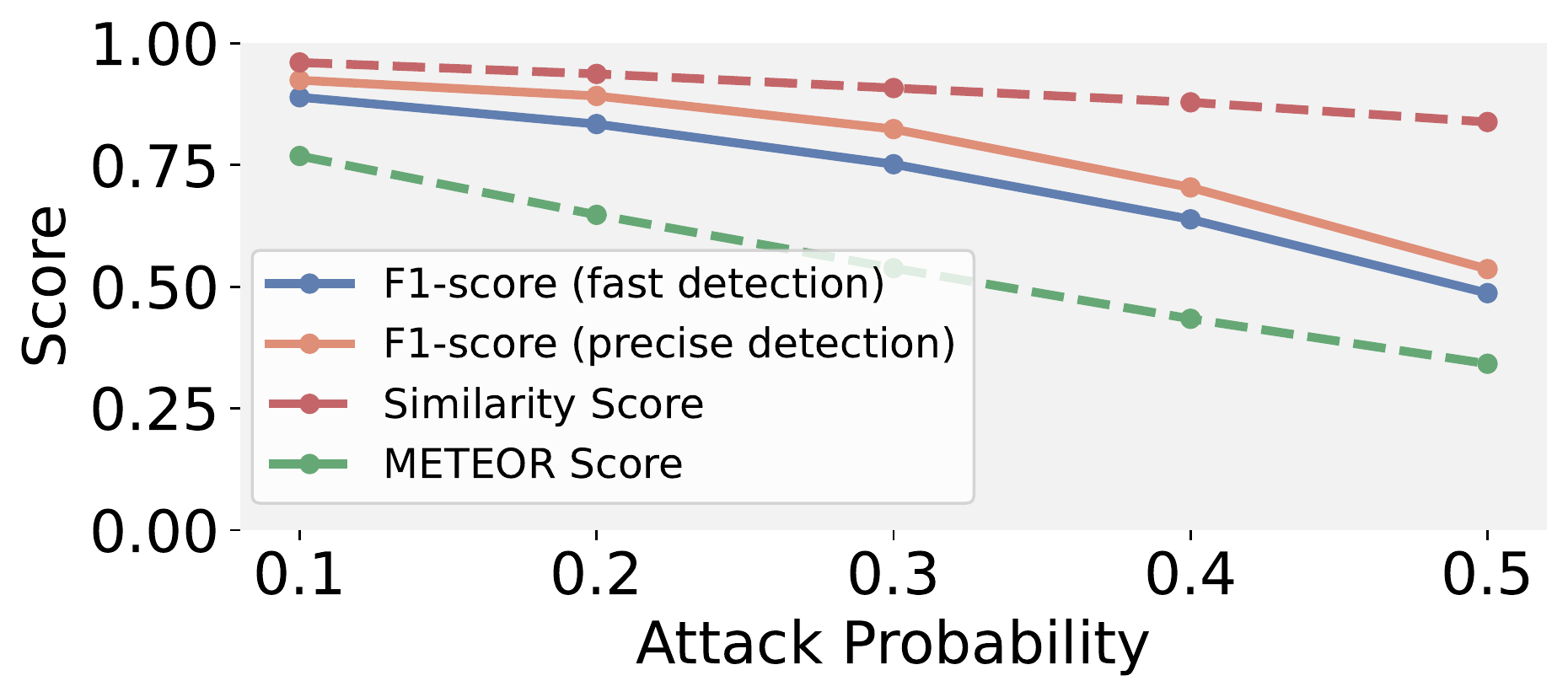}
        \caption{English}
    \end{subfigure}
    % \hspace{0.05\linewidth} % Adjust the horizontal space between the subfigures, if necessary
    \begin{subfigure}[b]{0.8\linewidth}
        \centering
        \includegraphics[width=\linewidth]{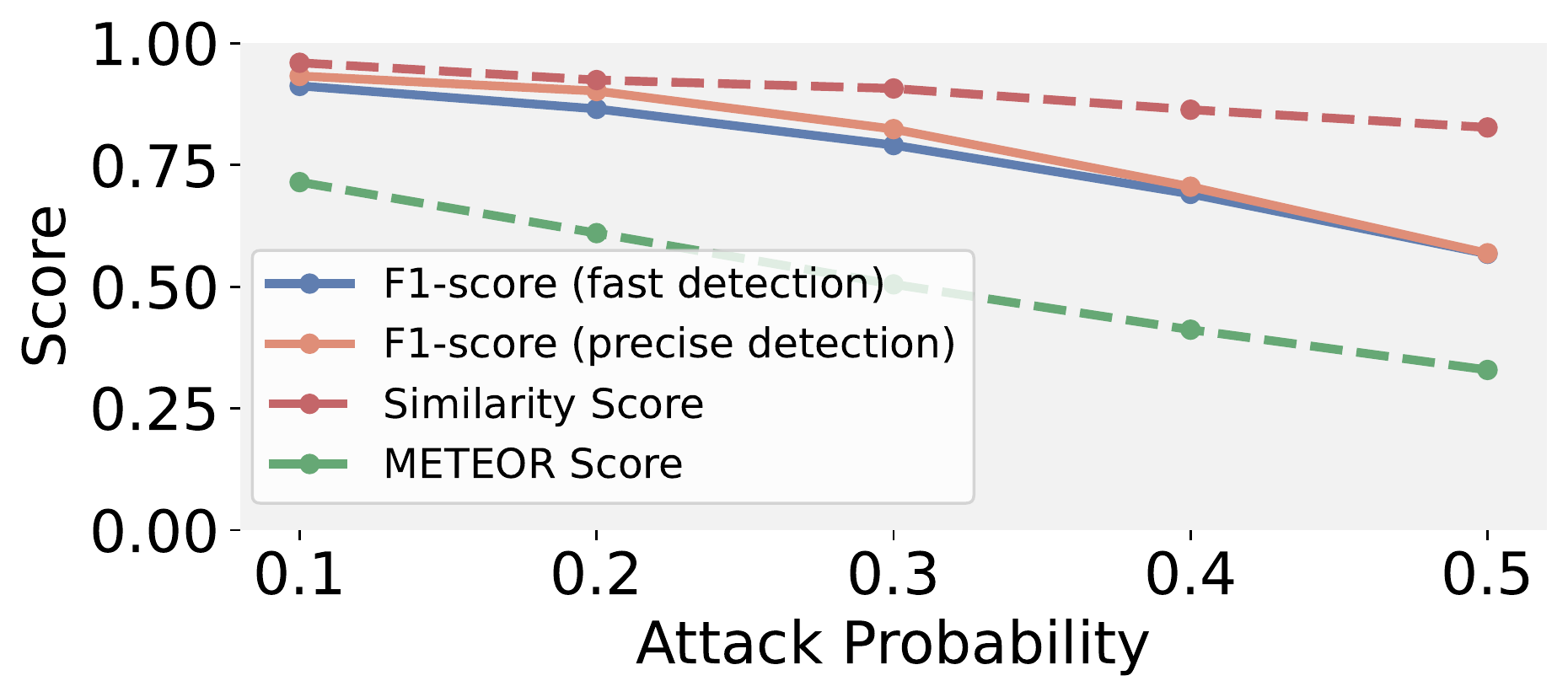}
        \caption{Chinese}
    \end{subfigure}
    \caption{Robustness analysis of the watermark under word deletion attacks.}
    \label{fig:remove_attack}
\end{figure}

\para{Word Deletion Attack}
To test the robustness of our watermark against word deletion attacks, we assign a deletion probability to each word (including symbols) and evaluate the watermark strength and text quality after the attack under different probabilities. As shown in Figure \ref{fig:remove_attack}, deletion attacks can severely damage the semantics of the text, causing the text quality to decline sharply as the attack probability (proportion of attacked content) increases. When the deletion probability of each word is 0.5, nearly half of the text content is deleted, making each sentence incomplete. As a result, our watermark is almost completely erased, and the attacked text also becomes unusable. However, when the removed content does not exceed 30\%, our watermark still maintains good detectability, even if the original semantics have been severely compromised.

\begin{figure}[t]
    \centering
    \begin{subfigure}[b]{0.8\linewidth}
        \centering
        \includegraphics[width=\linewidth]{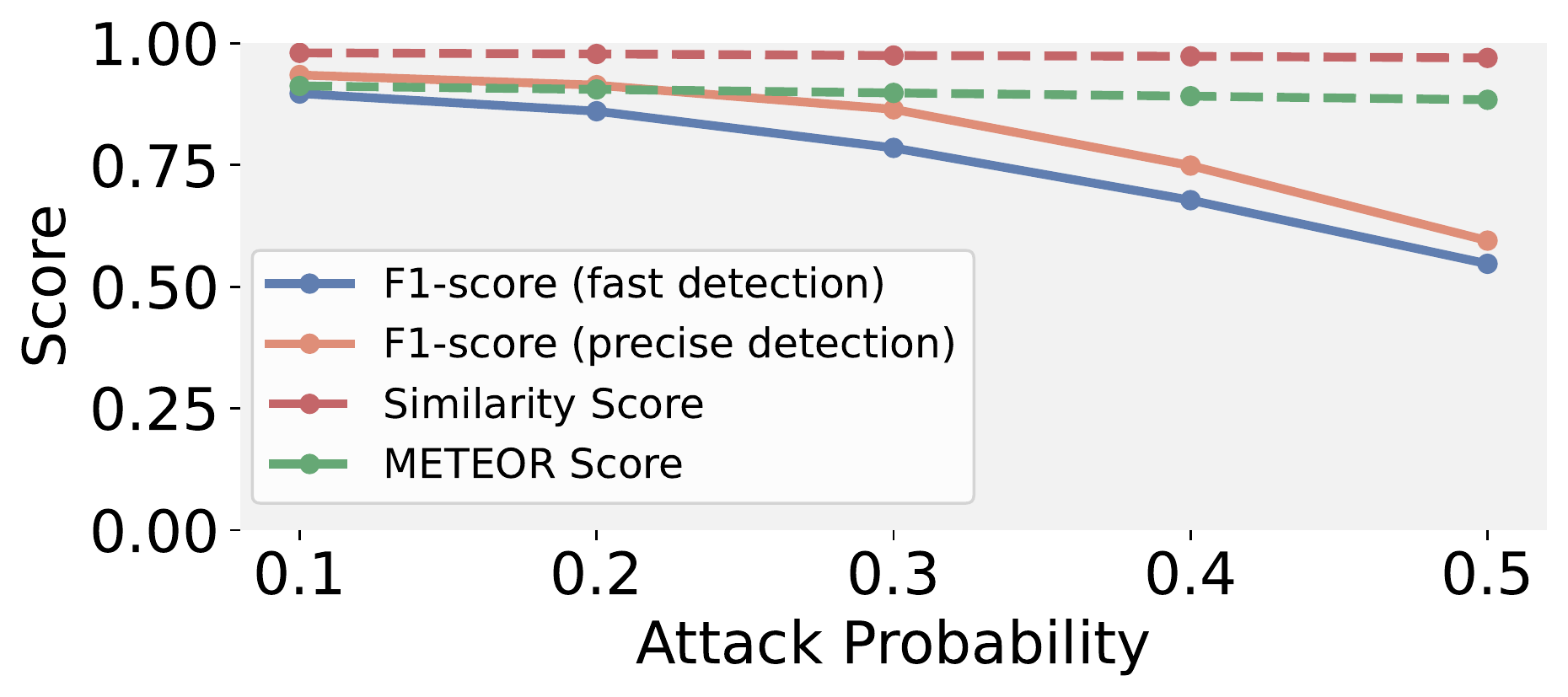}
        \caption{English}
    \end{subfigure}
    \begin{subfigure}[b]{0.8\linewidth}
        \centering
        \includegraphics[width=\linewidth]{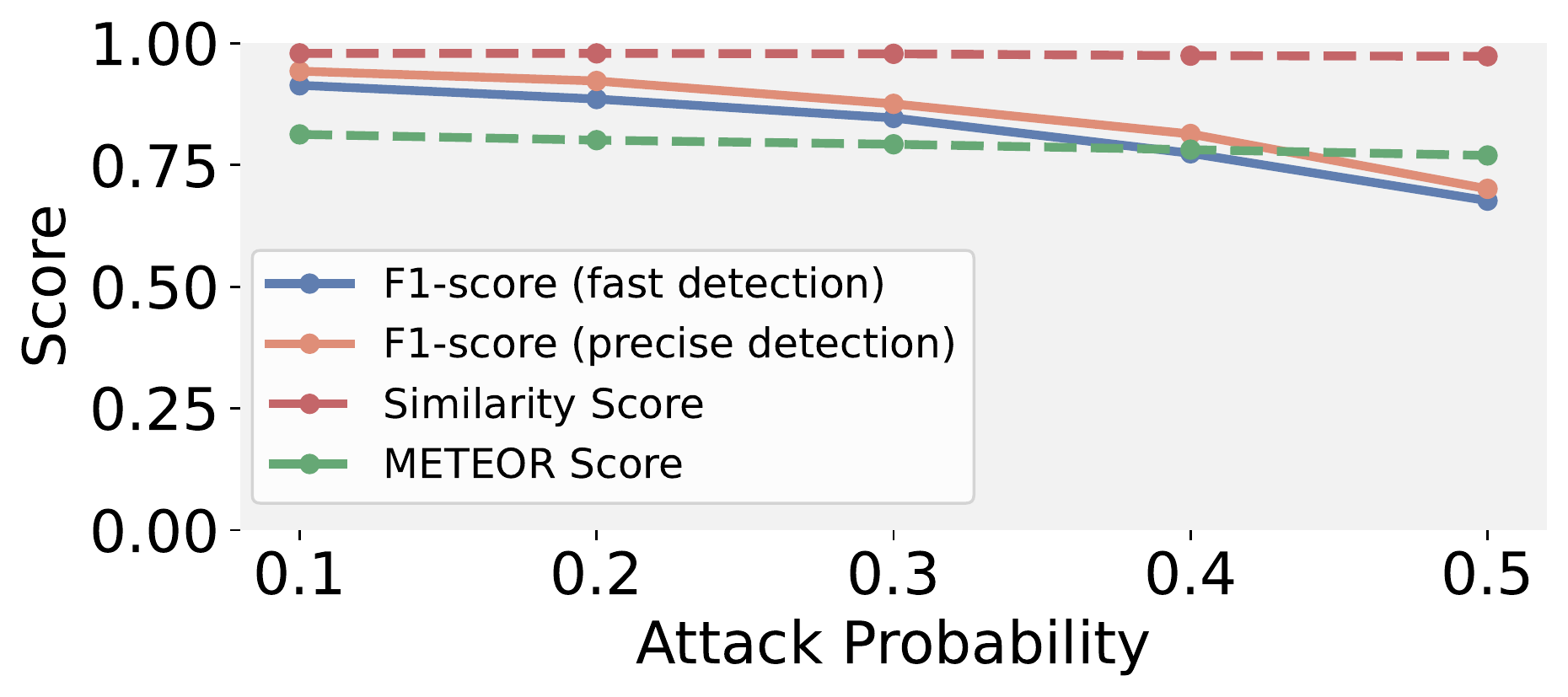}
        \caption{Chinese}
    \end{subfigure}
    \caption{Robustness analysis of the watermark under synonym substitution attacks.}
    \label{fig:synonym_attack}
\end{figure}
\begin{figure*}[t]
    \centering
    \begin{subfigure}{.25\textwidth}
        \centering
        \includegraphics[width=\linewidth]{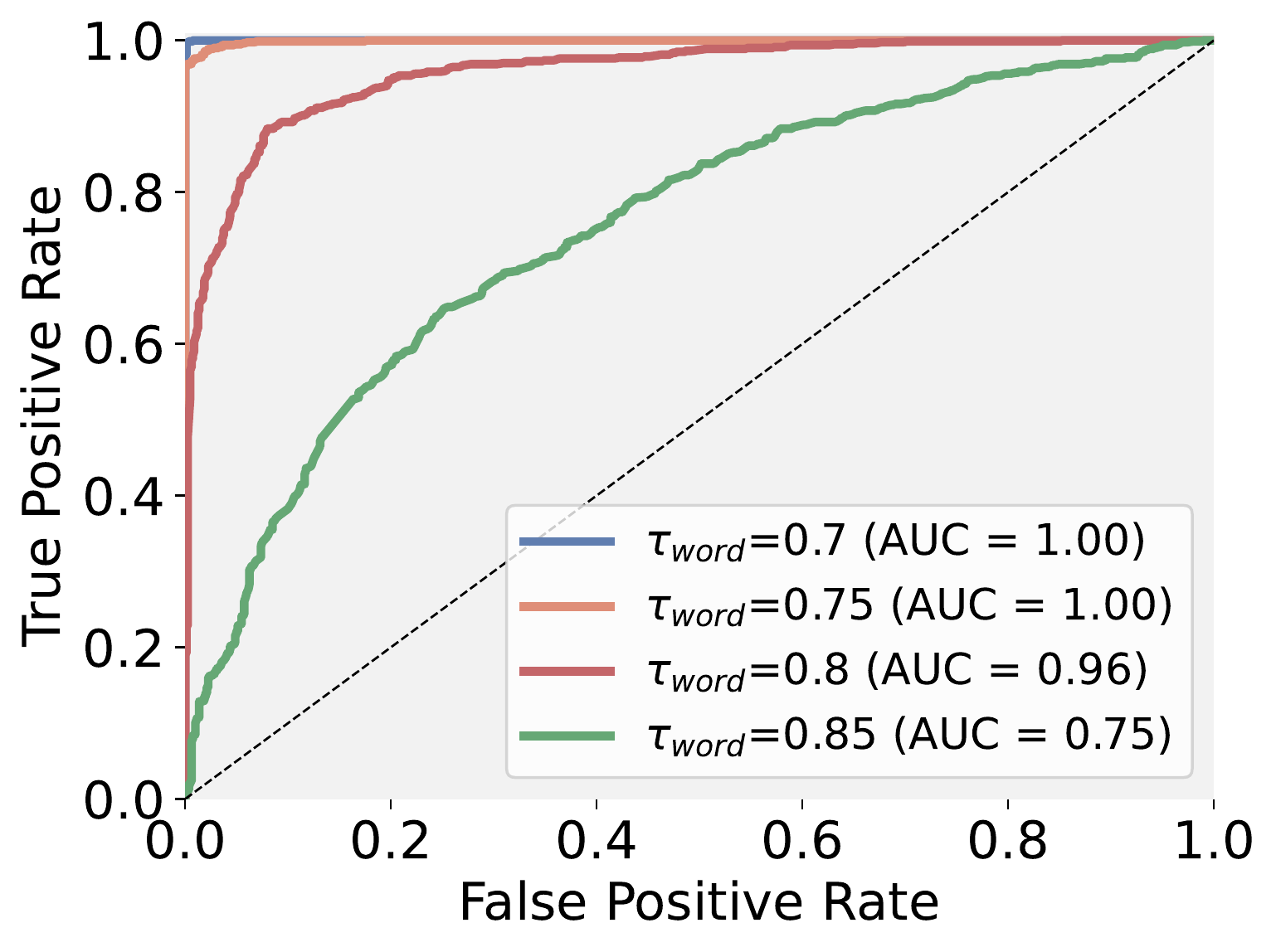}
        \caption{English \& Fast Detection}
        \label{fig:sub1}
    \end{subfigure}%
    \begin{subfigure}{.25\textwidth}
        \centering
        \includegraphics[width=\linewidth]{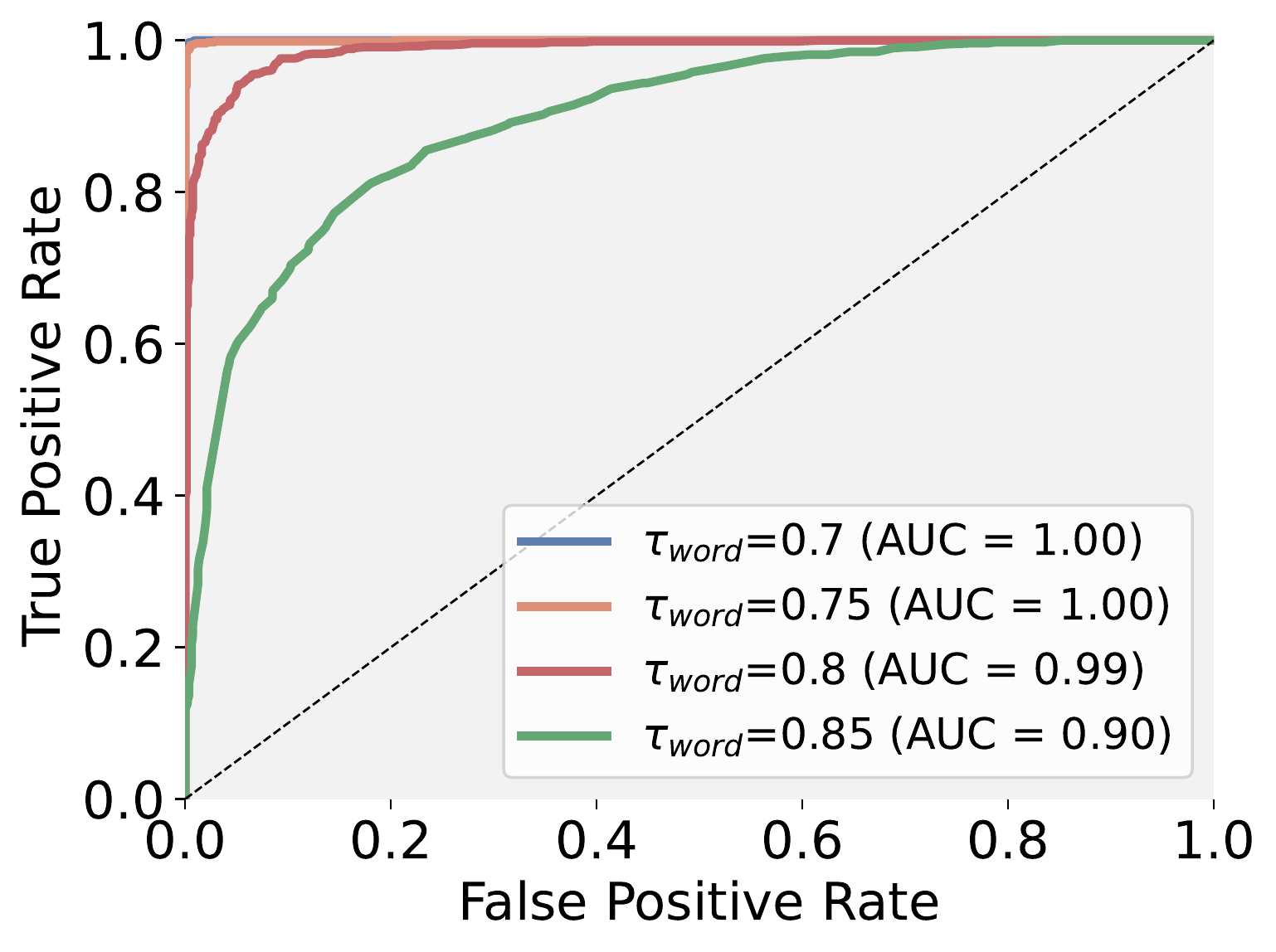}
        \caption{English \& Precise Detection}
        \label{fig:sub2}
    \end{subfigure}%
    \begin{subfigure}{.25\textwidth}
        \centering
        \includegraphics[width=\linewidth]{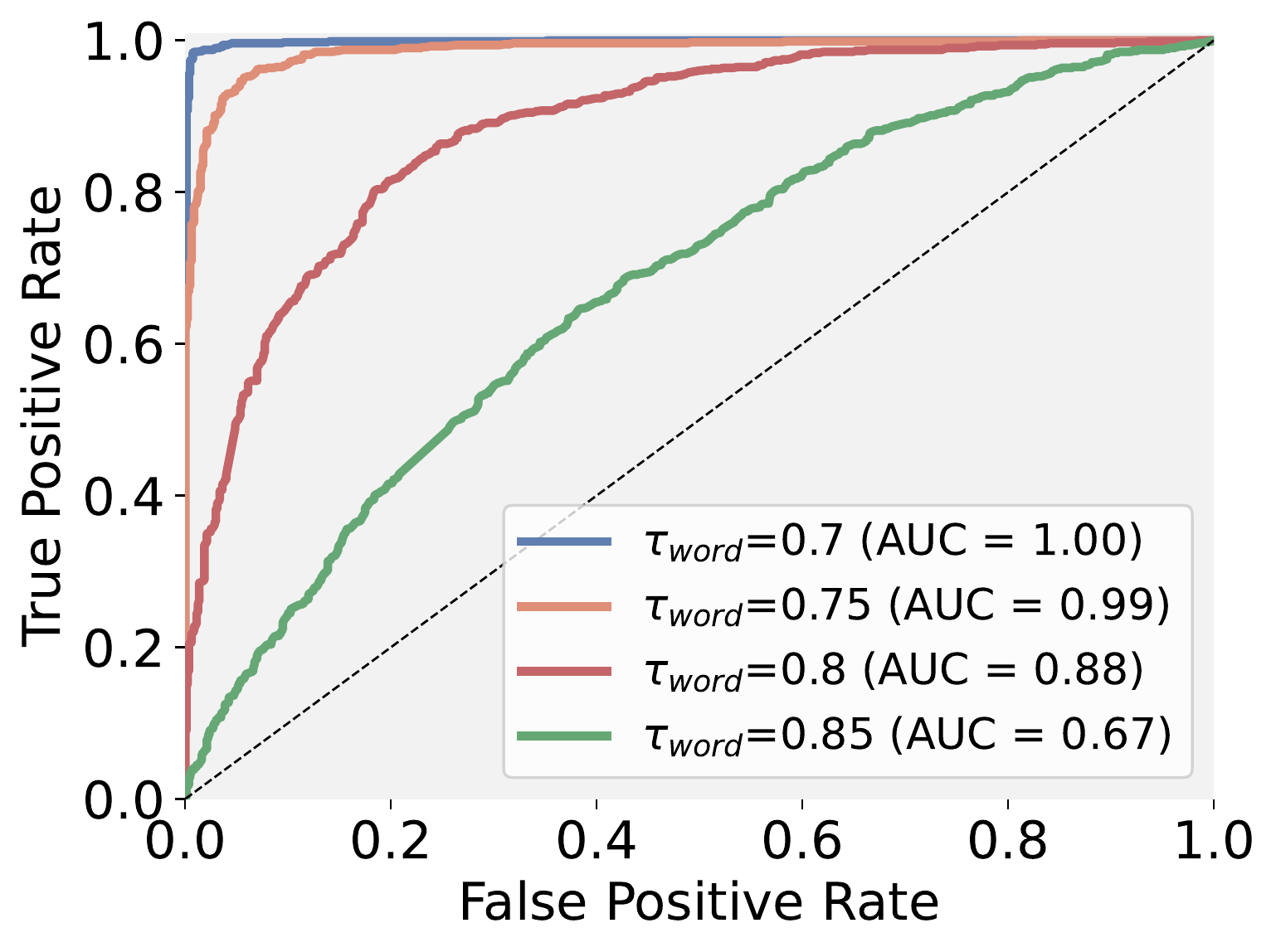}
        \caption{Chinese \& Fast Detection}
        \label{fig:sub3}
    \end{subfigure}%
    \begin{subfigure}{.25\textwidth}
        \centering
        \includegraphics[width=\linewidth]{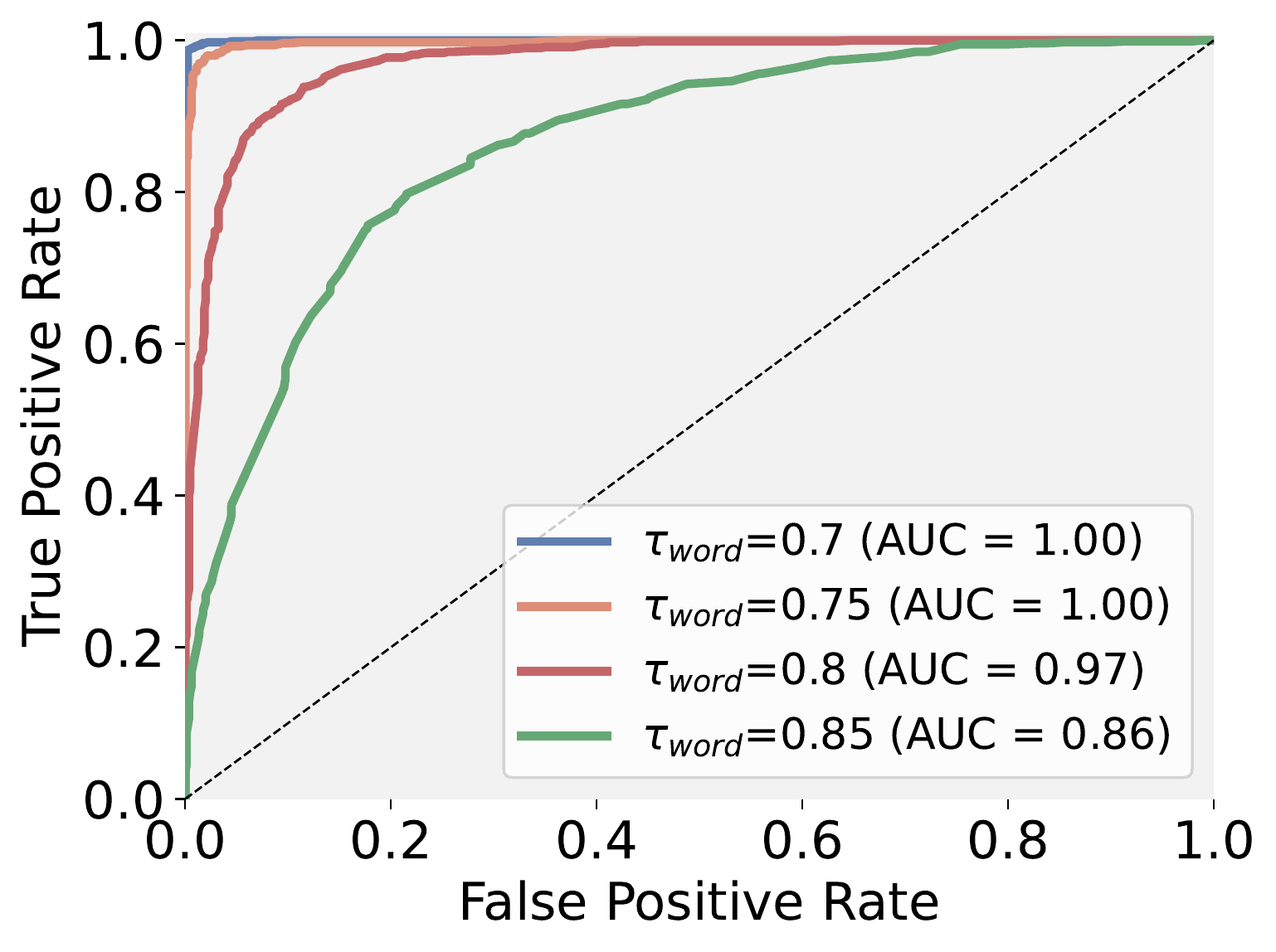}
        \caption{Chinese \& Precise Detection}
        \label{fig:sub4}
    \end{subfigure}
    \caption{ROC curves with AUC values for watermark detection in human-written text under different languages, $\tau_{\text{word}}$ values, and detection modes.}
    \label{fig:human_data}
\end{figure*}

\begin{figure}[t]
    \centering
    \includegraphics[width=0.9\linewidth]{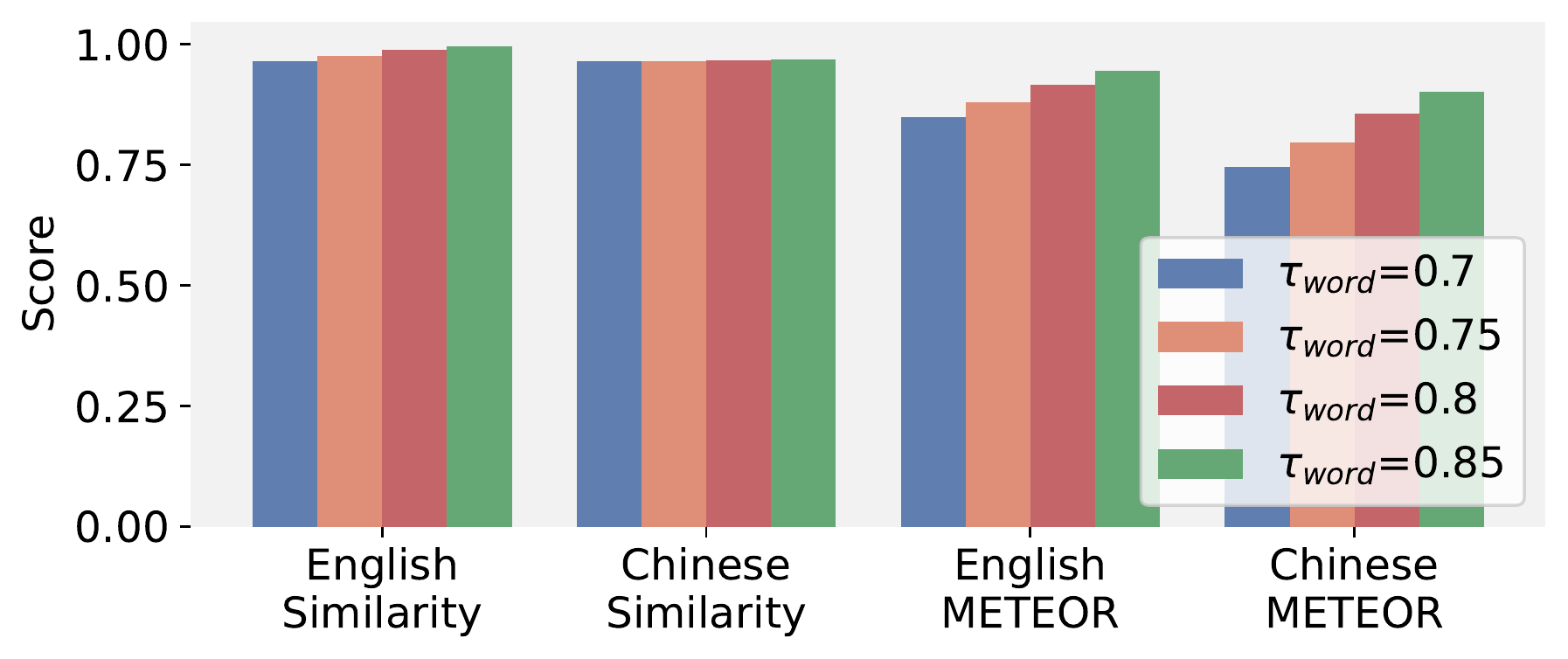}
    \caption{Semantic similarity and METEOR scores of watermarked human text compared to the original text under different $\tau_{\text{word}}$ values.}
    \label{fig:quality_human}
\end{figure}
\para{Synonym Substitution Attack}
Synonym substitution can happen when users are unhappy with certain word choices and modify them, or when attackers try to remove the watermark by changing words. To perform synonym substitution attack while preserving text quality as much as possible, we use our own synonym generation algorithm from \cref{ssec:injection} and also do not modify words that are unlikely to have high-quality synonyms. We set the probability of each word being replaced by a synonym and evaluate the watermark strength after the attack under different probabilities. 
% 对被动检测的影响
% To facilitate understanding, we present the results using F1-score, calculated with the significance level $\alpha=0.05$. 
As shown in Figure \ref{fig:synonym_attack}, the F1-scores decrease as the probability of synonym substitution increases. When each word is replaced with a 0.5 probability, our watermark is almost completely erased. This is because the synonym substitution process can be seen as a watermark rewriting attack on our watermarked text, and when half of the words are changed, all the binary encodings represented by each word and its preceding word are rewritten.

\yx{The results under sentence-level and word-level attacks illustrate that our watermark can be effectively preserved as long as the majority of watermark-bearing words are not modified.} In other words, our watermark can be linked to the semantics, making it hard for attackers to completely remove the watermark while maintaining the original text quality.
% However, it is important to note that if too many words are substituted, it may lead to a certain degree of damage to the original content, rendering the attack counterproductive.
\subsection{Time Cost of Fast and Precise Detection}
Table \ref{tab:time_cost} shows the average detection time per sample in both the fast and precise detection modes. A sample of about 200 words or 200 Chinese characters takes around 0.01 seconds to detect in fast mode. In precise detection, the average detection time is 7.646 seconds for an English sample of 200 words and 5.094 seconds for a Chinese sample of 200 characters, which has slightly less content than the English samples. This indicates that the precise detection spends more time analyzing text to enhance the detection capability.
\begin{table}[t]
\centering
\caption{Average detection time (seconds/sample) for English and Chinese texts in both fast and precise detection modes. We utilize a single NVIDIA GeForce RTX 2080 Ti GPU in the precise detection.}
\renewcommand{\arraystretch}{1}
\setlength{\tabcolsep}{2.8mm}{
\begin{tabular}{cccc}
\toprule[1.5pt]
Runtime & Fast Mode & Precise Mode \\ \hline
English              & 0.009     & 7.646        \\
Chinese               & 0.011     & 5.094        \\ 
\bottomrule[1.5pt]
\end{tabular}}
\label{tab:time_cost}
\end{table}

\subsection{Performance on Human Texts}
Previous experiments are conducted on generated text. Since the human brain's language system can currently also be considered a black box model, we expand our investigation to encompass human-written text, evaluating watermark strength under various $\tau_\text{word}$ values, following the experimental setup in \cref{ssec:Watermark Strength}. We use the English \texttt{FakeNews}\footnote{\url{https://www.kaggle.com/datasets/clmentbisaillon/fake-and-real-news-dataset}} and Chinese \texttt{People's Daily}\footnote{\url{https://www.kaggle.com/datasets/concyclics/renmindaily}} datasets, respectively. As shown in Figure \ref{fig:human_data}, watermark strength in human-written text is similar to that observed in generated text, as presented in Figure \ref{fig:tau}. Moreover, Figure \ref{fig:quality_human} indicates that the watermarked human-written text can maintain the original semantics. Therefore, our method is not only applicable to generated text but also to human-written text, thereby broadening its potential scope and utility.

\subsection{Comparison with Traditional Multi-bit Text Watermarking Methods}
\begin{table}[tbp]
      \centering
      \caption{Comparison of watermarking methods under word deletion attacks with different word deletion probabilities. We utilize $Z$-score ($\uparrow$) to measure the watermark strength.}
        \begin{tabular}{ccccc}
        \toprule[1.5pt]
        \multicolumn{1}{c}{Attack} & \multicolumn{2}{c}{Ours} & \multicolumn{1}{c}{\multirow{2}[3]{*}{Yang \etal \cite{yangxi}}} & \multicolumn{1}{c}{\multirow{2}[3]{*}{AWT \cite{awt}}} \\ 
\cmidrule(r){2-3}     \multicolumn{1}{c}{Probability}  & \multicolumn{1}{c}{Fast} & \multicolumn{1}{c}{Precise} &    &  \\
\hline
        0.1 & \textbf{3.00}  & \textbf{3.63} & 2.19 & 2.73 \\
        0.2 & \textbf{2.50} & \textbf{2.96} & 1.42 & 2.46 \\
        \bottomrule[1.5pt]
        \end{tabular}%
      \label{tab:compare_remove}%
    \end{table}%
Traditional multi-bit text watermarking methods, as discussed in \cref{ssec:multibit_watermark}, focus on embedding multiple bits of information for copyright protection and leak tracing, which makes them difficult to compare with our method directly. \yx{
Therefore, we restructure these methods to achieve the functionality of watermark detection. Specifically, we set the embedded multi-bit watermark sequence to consist of repeated bit-1s and employ the hypothesis test in our method to examine the extracted bit sequence for watermark detection.} Then, we use the word deletion attack for robustness comparison (setting the deletion probability to only 0.1 and 0.2, which is sufficient to demonstrate the differences between these methods). As shown in Table \ref{tab:compare_remove}, the method proposed by Yang \etal \cite{yangxi} exhibits sensitivity to contextual changes, resulting in a rapid decline in watermark detection performance when a small portion of words is deleted. Although AWT \cite{awt} can provide stronger robustness, it significantly sacrifices semantic quality by introducing words or symbols that disrupt the semantic structure of the text, as shown in Table \ref{tab:compare_awt}. However, this severely compromises fidelity. In contrast, our method strikes a balance between robustness and fidelity, maintaining the original semantics without introducing grammatical errors while also providing better watermark strength.

\begin{table}[t]
    \newcolumntype{L}{>{\arraybackslash}m{3.3cm}}
    \renewcommand{\arraystretch}{1.5}
    \centering
    \caption{Examples of watermarked sentences compared with AWT. The substituted words are underlined.} \label{tab:compare_awt}
    \resizebox{1\linewidth}{!}{%
        \begin{tabular}{L|L|L}
            \toprule[1.3pt]
				\textbf{Original} & \textbf{AWT} \cite{awt} & \textbf{Ours} \\\midrule
				resulting in a population decline as workers left for other areas & resulting in a population decline \ul{an} workers left for other areas & resulting in a population \ul{dip} as workers left for other areas \\
				but the complex is broken up by the heat of cooking & \ul{and} the complex is broken up by the heat of cooking & but the complex is \ul{torn} up by the heat of cooking
                \\
				For the second part of the show, Carey had the second costume change of the evening, donning a long $<$unk$>$ black gown and semi @-@ $<$unk$>$ hair. & For the second part of the show, Carey had the second \ul{Buddhist} change of the evening, \ul{were} a long $<$unk$>$ black gown and semi @-@ $<$unk$>$ hair.  & For the second part of the \ul{program}, Carey had the second costume change of the \ul{night}, donning a \ul{lengthy} $<$unk$>$ black gown and semi @-@ $<$unk$>$ hair.
				\\\bottomrule[1.3pt]
    \end{tabular}}
% 		\vspace{-4mm}
\end{table}
\begin{table}[t]
    \small
      \centering
      \caption{Ablation study results showcasing the importance of sentence-level and word-level semantic similarity constraints in maintaining the semantic quality of the watermarked text during watermark injection.}
        \begin{tabular}{>{\centering\arraybackslash}p{5em}p{23.875em}}
        \toprule[1.5pt]
        \multirow{3}{*}{Original} & 
        In the warm embrace of the golden sun, I stroll through the vibrant garden, filled with the delightful aroma of blossoming flowers. 
        % A gentle breeze whispered sweet melodies, as the joyful birds danced in the sky above. 
        The lush green grass caressed my feet,  ...
        \\ \hline
       \multirow{3}{*}{Watermarked} & In the warm \ul{hug} of the golden sun, I \ul{walk} through the \ul{brilliant} garden, \ul{full} with the delightful aroma of blossoming flowers. 
        % A gentle breeze whispered lovely melodies, as the joyful birds danced in the sky above. 
        The lush green \ul{{{lawn}}} caressed my feet, ...
        \vspace{0.5pt}
        \\ \hline
        \multirow{3}{5em}{Watermarked w/o $\tau_\text{sent}$} & In the \ul{cool} \ul{{hug}} of the golden sun, I \ul{walk} through the \ul{brilliant} garden, filled with the delightful aroma of blossoming flowers. 
        % A gentle breeze whispered lovely melodies, as the joyful birds danced in the sky above. 
        The lush \ul{purple} \ul{soil} caressed my feet, ...
        \vspace{0.5pt}
        \\ \hline
        \multirow{3}{5em}{Watermarked w/o $\tau_\text{word}$} & In the \ul{light} embrace of the golden \ul{sunshine}, I \ul{walk} through the \ul{{flower}} \ul{yard}, \ul{{full}} with the delightful aroma of blossoming \ul{{petal}}. 
        % A gentle breeze sing soft tune, as the joyful birds play in the sun above. 
        The lush \ul{{grass}} grass caressed my \ul{shoe}, ...
        \\
        \bottomrule[1.5pt]
        \end{tabular}%
      \label{tab:ablation}%
    \end{table}%

\subsection{Ablation: Assessing the Roles of $\tau_\text{sent}$ and $\tau_\text{word}$ in Semantic Quality Control}
We conduct ablation experiments on the semantic quality control modules to demonstrate the necessity of each component. The semantic control module comprises two parts: the sentence-level semantic similarity and the word-level semantic similarity. The sentence-level constraint requires the similarity score between the watermarked and original text to exceed $\tau_\text{sent}$. The word-level constraint requires the weighted average of global and contextual similarity between the word used for watermarking and the original word to exceed $\tau_\text{word}$. Table \ref{tab:ablation} shows that, without the sentence-level constraint, the original words may be replaced by statistically similar words that express different or opposite semantics (\eg `warm'-`cool', `green'
-`purple', and `grass'-`soil'). In the absence of the word-level constraint, low-quality, irrelevant, or grammatically erroneous words are introduced (\eg `green grass'-`grass grass'). Only with $\tau_\text{sent}$ and $\tau_\text{word}$ together can we ensure the semantic quality of the watermarked text.

\section{Discussion} \label{sec:discussion} 
\para{Comparison with the Watermarking Method for White-Box Language Models}
The watermarking method proposed by Kirchenbauer \etal \cite{maryland} is intended for model owners and requires the control to the model's output probability distribution. This characteristic makes it infeasible in black-box language model usage scenarios where the probability distribution information is not available.
% Our method, designed for black-box scenarios, injects watermarks by replacing synonyms in the generated text. 
Unlike the white-box method that generates watermarked text from scratch, our method modifies the already generated text and can maintain the original semantics, form, and style. 
We contend that these two methods are more complementary than competitive. 
If needed, black-box watermarking, white-box watermarking, and passive detection techniques can be combined to offer multiple detection results for achieving high robustness. OpenAI also pointed out that the company was researching watermarking as a form of detection, and that it could complement the passive detection tool.
\para{Reasons to Use BERT for Synonym Generation}
The use of BERT models is attributed to their superior synonym generation performance among open-source models. It should be emphasized that our main goal is to propose a general framework for watermarking text generated by black-box language models. The specific algorithms within the framework are adaptable. As more efficient synonym generation algorithms emerge in the future, they can be readily incorporated into our framework.
% White-box watermarking methods interfere with the generation process, providing ample watermark embedding space as the same context can lead to multiple continuations. This results in watermark embedding on both form and style, meaning that watermarked generated text and non-watermarked generated text might be entirely different under the same prompt. In contrast, our method is based on the output of a black-box model and only replaces a few words or characters with synonyms, linking the watermark information to the semantic information of the text while maintaining the original semantics intact and consistent with the original text in terms of form and style.

% Since white-box watermarking is achieved by generating text from scratch, and our watermarking is achieved by modifying the already generated text, it is challenging to align the experimental settings to compare the two. In fact, we believe that the two methods are more complementary than competitive. If needed, they can be used in conjunction to implement a joint watermark, allowing both the API provider and third parties to perform detection or authentication.

\section{Conclusion}
In this paper, we propose a watermarking framework for injecting authentication watermarks into text generated from black-box language models. The motivation is to enable third-parties who employ black-box language model services (\eg APIs) to autonomously inject watermarks in their generated text for the purposes of detection and authentication. Extensive experiments on text datasets with different languages and topics (\textbf{Generality}) have demonstrated that the watermark retains a connection to the original semantics (\textbf{Fidelity}), making it challenging for adversaries to remove the watermark without affecting the integrity of the original content (\textbf{Robustness}). We hope our method can provide new insights for generated text detection and inspire more future work.

\message{^^JLASTBODYPAGE \thepage^^J}

\clearpage
\balance
\bibliographystyle{ACM-Reference-Format}

\balance

\bibliography{main}

\message{^^JLASTREFERENCESPAGE \thepage^^J}

\ifincludeappendixx
	\clearpage
	\appendix
	\section{Runtime Optimization}
In this paper, the watermark injection and detection algorithms are presented in a sequential iterative manner for selected words to facilitate easy understanding. However, in real-world engineering applications, this may lead to increased runtime. To expedite the watermark injection and detection process, we can implement parallel processing on multiple words in the given text, since the original text is known. We first compute the POS for each word in the text and record the index positions of words meeting the POS criteria in an index list. Then, we employ BERT to generate synonym candidates for all words with positions in the list. For each candidate, we substitute it in the corresponding position of the original text to create a new text variant. Once all candidate texts are generated, we utilize batch processing to compute the similarity scores for each candidate word simultaneously, significantly reducing time for both watermark injection and precise detection. Finally, we further refine the candidates based on their similarity scores and choose the best synonyms for watermark injection using our synonym sampling algorithm. Both the iterative and parallel algorithms will be included in the released code.

\section{Demo and Source Code}
\begin{figure*}[b]
    \centering
    \includegraphics[width=\linewidth]{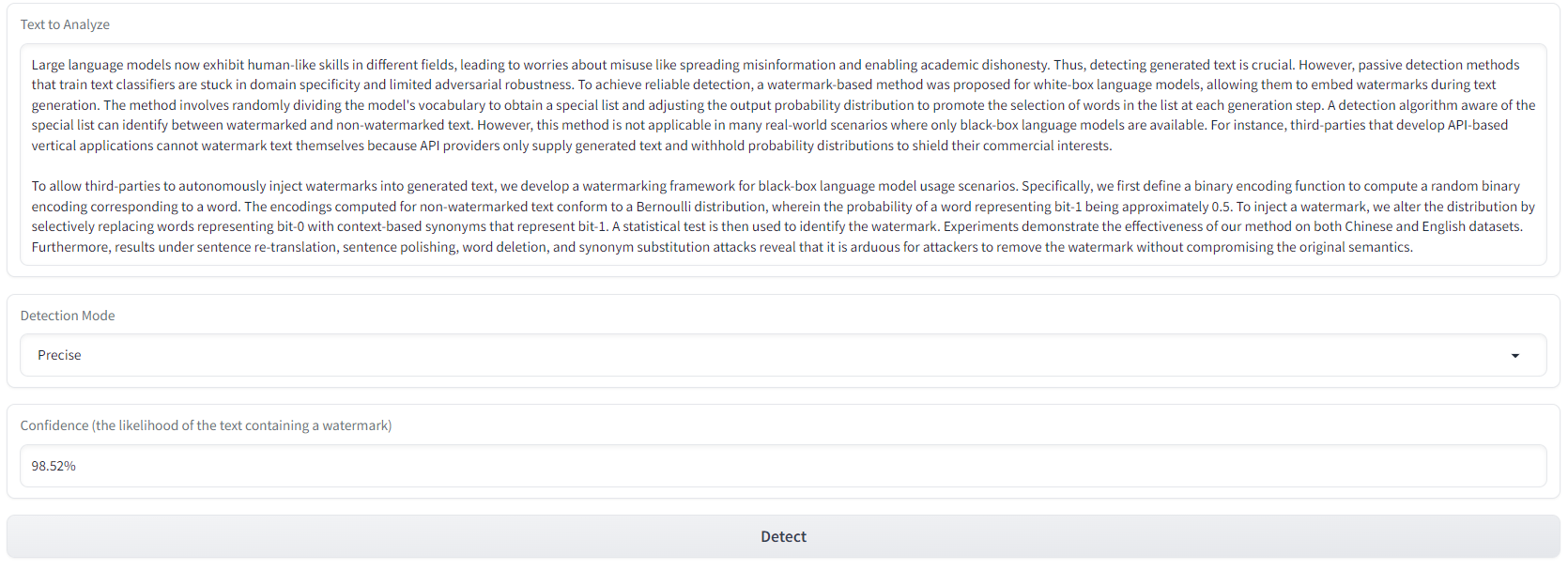}
    \caption{Screenshot of using our demo to perform watermark detection on the abstract of this paper.}
    \label{fig:abstract}
\end{figure*}
In the additional materials, we provide a demo for watermark injection and detection based on Gradio\footnote{\url{https://gradio.app/}}, including both the source code and a demonstration video. The watermark carried in the abstract can be detected by the detector in the demo, with a confidence level of 98.52\%, as shown in Figure \ref{fig:abstract}.

\section{Pseudocode for the Attacks} 
We provide further details related to the attacks used in our robustness analysis. The process of re-translation attack is illustrated in Algorithm \ref{alg:retrans}, where we utilize the commercial Baidu Translation API and DeepL API as translators. The process of polishing attack is illustrated in Algorithm \ref{alg:polish}, where we employ GPT-3.5-turbo API\footnote{\url{https://platform.openai.com/docs/models/gpt-3-5}} to perform sentence polishing. In the same manner, the pseudocode for word deletion and synonym substitution attacks can be found in Algorithm \ref{alg:deletion} and Algorithm \ref{alg:synonym}, respectively.

\section{More Examples}
In the additional materials, we provide text files (refer to \texttt{english\_sa\\mples.txt} and \texttt{chinese\_samples.txt}) containing the original and watermark texts utilized in our experiments, comprising 800 samples each for both Chinese and English. We set $\tau_{word}=0.8$ for English text and $\tau_{word}=0.75$ for Chinese text. 

\begin{figure*}[ht]
  \begin{minipage}{0.45\textwidth}
    \begin{algorithm}[H]
    \caption{Re-translation Attack}
    \label{alg:retrans}
    \renewcommand{\algorithmicrequire}{\textbf{Input:}}
    \renewcommand{\algorithmicensure}{\textbf{Output:}}
    \begin{algorithmic}[1]
    \Require $watermarked\_text$, the attack probability $p$
    % \Comment{$p$ repres}
    \Ensure $attacked\_text$
    \State $src \gets \text{"ENG" or "CN"}$
    \If{$src == \text{"ENG"}$}
        \State $inter \gets \text{"CN"}$
    \Else
        \State $inter \gets \text{"ENG"}$
    \EndIf
    \For{each $sent$ in $watermarked\_text$}
        \State $rand\_num \gets random(0, 1)$
        \If{$rand\_num <= p$}
            \State $trans\_sent \gets translator(sent, src, inter)$
            \State $retrans\_sent \gets translator(trans\_sent, inter, src)$
            % \State $sent \gets retrans\_sent$
        \EndIf
    \EndFor
    \State $attacked\_text \gets$ concatenated sentences
    \end{algorithmic}
    \end{algorithm}
  \end{minipage}
  \hspace{0.5cm}
  \begin{minipage}{0.5\textwidth}
    \begin{algorithm}[H]
\caption{Polishing Attack}
\label{alg:polish}
\renewcommand{\algorithmicrequire}{\textbf{Input:}}
\renewcommand{\algorithmicensure}{\textbf{Output:}}
\begin{algorithmic}[1]
\Require $watermarked\_text$, $prompt$, the attack probability $p$
\Ensure $attacked\_text$
\For{each $sent$ in $watermarked\_text$}
    \State $rand\_num \gets random(0, 1)$
    \If{$rand\_num <= p$}
        \State $polished\_sent \gets \texttt{GPT-3.5-turbo}(prompt,sent)$
        % \State $sent \gets polished\_sent$
    \EndIf
\EndFor
\State $attacked\_text \gets$ concatenated sentences
\end{algorithmic}
\end{algorithm}
  \end{minipage}
\end{figure*}

\begin{figure*}[ht]
  \begin{minipage}{0.45\textwidth}
    \begin{algorithm}[H]
\caption{Word Deletion Attack}
\label{alg:deletion}
\renewcommand{\algorithmicrequire}{\textbf{Input:}}
\renewcommand{\algorithmicensure}{\textbf{Output:}}
\begin{algorithmic}[1]
\Require $watermarked\_text$, the attack probability $p$
\Ensure $attacked\_text$
\State $sentences \gets$ split $watermarked\_text$ into sentences
\For{each $sent$ in $sentences$}
    \State $words \gets$ split $sent$ into words (including symbols)
    \For{each $word$ in $words$}
        \State $rand\_num \gets random(0, 1)$
        \If{$rand\_num <= p$}
            \State remove $word$ from $sent$
        \EndIf
    \EndFor
\EndFor
\State $attacked\_text \gets$ concatenated words
\end{algorithmic}
\end{algorithm}
  \end{minipage}
  \hspace{0.5cm}
  \begin{minipage}{0.5\textwidth}
    \begin{algorithm}[H]
\caption{Synonym Substitution Attack}
\label{alg:synonym}
\renewcommand{\algorithmicrequire}{\textbf{Input:}}
\renewcommand{\algorithmicensure}{\textbf{Output:}}
\begin{algorithmic}[1]
\Require $watermarked\_text$, $p$
\Ensure $attacked\_text$
\State $sentences \gets$ split $watermarked\_text$ into sentences
\For{each $sent$ in $sentences$}
    \State $words \gets$ split $sent$ into words
    \For{each $word$ in $words$}
        \If{POSFilter($word$)} 
        \Comment{\cref{alg:watermark_injection}}
            \State $rand\_num \gets random(0, 1)$
            \If{$rand\_num <= p$}
                \State $C \gets$ SynonymsGen($sent$, $word$) \Comment{\cref{alg:watermark_injection}}
                \State $C' \gets$ FilterCandidates($sent$, $C$, $word$) \Comment{\cref{alg:watermark_injection}}
                \If{$C' \neq \emptyset$}
                    \State Substitute $word$ with the first word in $C'$
                \EndIf
            \EndIf
        \EndIf
    \EndFor
\EndFor
\State $attacked\_text \gets$ concatenated sentences
\end{algorithmic}
\end{algorithm}
  \end{minipage}
\end{figure*}

% \begin{figure*}[ht]
%   \begin{minipage}{0.45\textwidth}
%     \begin{algorithm}[H]
%       \caption{Word Deletion Attack}
%       \label{alg:deletion}
%       % ... algorithm content ...
%     \end{algorithm}
%   \end{minipage}
%   \hfill
%   \begin{minipage}{0.45\textwidth}
%     \begin{algorithm}[H]
%       \caption{Synonym Substitution Attack}
%       \label{alg:synonym}
%       % ... algorithm content ...
%     \end{algorithm}
%   \end{minipage}
% \end{figure*}

\fi

\message{^^JLASTPAGE \thepage^^J}

\end{document}